\documentclass[sigconf, nonacm]{aamas} 
\usepackage{balance} 
\usepackage{xspace} 
\usepackage{tabularx}
\usepackage{booktabs}
\usepackage{longtable}
\usepackage{pdflscape}
\usepackage{array}
\usepackage{url}
\usepackage{multirow}
\usepackage{hyperref}
\usepackage{enumitem}

\usepackage{xcolor}
\usepackage{tcolorbox}

\tcbuselibrary{breakable}

\tcbset{
    takeawaystyle/.style={
        sharp corners,
        colback = white,
        before skip= 3pt,
        after skip= 3pt,
    },
    promptstyle/.style={
        breakable,
        sharp corners,
        boxrule=0.4pt,
        leftrule=2pt,
        left=4pt,
        right=4pt,
        top=4pt,
        bottom=4pt,
        fontupper=\footnotesize,
        fonttitle=\footnotesize\bfseries,
        coltitle=black,
        before skip=3pt,
        after skip=3pt,
        left=1pt,
        right=1pt,
        top=1pt,
        bottom=1pt
    }
}

\newtcolorbox{boxH}{
    takeawaystyle,
    colback = gray!8, 
    colframe = black!60,
    boxrule = 0.5pt, 
    leftrule = 3pt,
    left=1pt,
    right=1pt,
    top=1pt,
    bottom=1pt
}

\newtcolorbox{hclprompt}[1]{
    promptstyle,
    colback=orange!4!white,
    colframe=orange!60!black,
    colbacktitle=orange!10!white,
    title={#1}
}

\newtcolorbox{hllprompt}[1]{
    promptstyle,
    colback=teal!4!white,
    colframe=teal!60!black,
    colbacktitle=teal!10!white,
    title={#1}
}

\newcommand{\name}{\texttt{AgBench}\xspace}

\makeatletter
\gdef\@copyrightpermission{
  \begin{minipage}{0.2\columnwidth}
   \href{https://creativecommons.org/licenses/by/4.0/}{\includegraphics[width=0.90\textwidth]{by}}
  \end{minipage}\hfill
  \begin{minipage}{0.8\columnwidth}
   \href{https://creativecommons.org/licenses/by/4.0/}{This work is licensed under a Creative Commons Attribution International 4.0 License.}
  \end{minipage}
  \vspace{5pt}
}
\makeatother

\setcopyright{none}

\title{
AgBench: Agentic AI Benchmarks for Personal AI Devices
}

\author{Yizhou Han}
\affiliation{
  \institution{University of St Andrews \country{UK}}
  }
\email{yh91@st-andrews.ac.uk}

\author{Di Wu}
\affiliation{
  \institution{Zhejiang University of Technology \country{China}}
  }
\email{datawonder8@gmail.com} 

\author{Dhananjay Saikumar}
\affiliation{
  \institution{University of St Andrews \country{UK}}
  }
\email{ds304@st-andrews.ac.uk} 

\author{Blesson Varghese}
\affiliation{
  \institution{University of St Andrews \country{UK}}
  }
\email{bv6@st-andrews.ac.uk} 

\begin{abstract}
Agentic AI systems increasingly rely on cloud-hosted large language models for planning, tool use, and iterative execution, raising concerns about API cost and data exposure. Advances in personal AI devices enable agents to execute locally, but limited resources on device may affect task success and performance. Existing benchmarks are inadequate for systematically characterizing these trade-offs across devices, workloads, and deployment architectures. We present \name, a benchmark suite and open artifacts for reproducible evaluation of agentic AI on personal devices. Using \name, we evaluate local, hybrid, and cloud execution across agentic workloads, examining task success, latency, cloud API cost, and data exposure. 
Our results, drawn from over $162.07$ million data points, show that personal AI devices can complete many agent tasks locally, but local-only execution generally has lower task success and longer completion times than cloud-only execution, especially as concurrency increases. Local-only execution eliminates cloud model API costs and sensitive-information exposure to cloud agents. Hybrid execution can improve task success, but its cloud cost and data exposure depend on how agents divide work and share information. No single architecture performs best across task success, goodput, cloud cost, and data exposure; deployment choices should reflect the intended workload and device capabilities.
\name is available at \url{https://anonymous.4open.science/r/AgBench-2777}.
\end{abstract}


\newcommand{\BibTeX}{\rm B\kern-.05em{\sc i\kern-.025em b}\kern-.08em\TeX}

\begin{document}




\maketitle 
\pagestyle{plain}
\thispagestyle{plain}

\section{Introduction}
\label{sec:introduction}
Agentic AI is an emerging area in which systems comprising
one or more agents built on foundation models pursue user-defined goals by planning and iteratively using tools, evaluating feedback, and refining actions~\cite{wang2025openhands,wu2026agenticedgeai, hsiao2026procedural, choi2026reactree}. Such systems are seen in software engineering, workplace productivity tools, and personal assistance applications~\cite{agashe2024agent-s, wang2026openhandsdk,zhang2025ufo2,zhou2026mobile-rag}. 
Recent research found that software coding agents were adopted by up to 28.7\% of active GitHub projects, and a 15-fold year-over-year increase in active agents was reported across the Microsoft 365 ecosystem in 2026~\cite{robbes2026agentic,microsoft2026worktrend}.

Agent systems, such as Claude Code and Codex, rely on cloud-hosted large language models (LLMs) for planning, evaluating intermediate results, and refining subsequent actions~\cite{anthropic_claude_code, bolin2026codex}. This requires user inputs and the execution context to be transferred to cloud services, raising privacy concerns. Additionally, repeated model calls in long-running workflows can incur substantial API costs~\cite{yang2024swe}. These constraints, coupled with advances in hardware with dedicated AI accelerators and large unified memory, have fostered growing interest in running agents on \emph{personal AI devices}, which are user-controlled devices with sufficient compute and memory to run AI agents locally~\cite{nvidia_dgx_spark, amd_agent_cp}. Representative devices include RTX 5090-based PCs, NVIDIA DGX Spark, and AMD Ryzen AI Halo.

Moving agent execution from the cloud to personal AI devices introduces a fundamental trade-off. Local execution reduces cloud API costs and local data exposure to the cloud, but the relative resource constraints limit the size of models that can be used, which in turn impacts execution performance, reduces task success and increases execution latency. Hybrid architectures, which sit between fully local and fully cloud deployments, can distribute agent components across the device and cloud, potentially balancing competing objectives. However, their benefits remain unclear.
Existing benchmarks are inadequate for systematically characterizing how trade-offs vary across devices, workloads, and deployment architectures. 

This leads to a \textbf{central question}: \emph{Are personal AI devices ready for agentic AI?} We address this by considering four research questions: 

\textbf{Q1:} What is the impact on task success when agents move from the cloud to personal AI devices? 

\textbf{Q2:} Does local execution slow agent workflows? 

\textbf{Q3:} Can local execution reduce cloud API costs? 

\textbf{Q4:} How much cloud data exposure can local execution reduce? 

Beyond these four research questions, we further examine when local, hybrid, and cloud execution offer the best overall trade-offs.

Our study reveals four key findings.
First, task success varies across workloads and deployment architectures, with local-only execution falling further behind cloud-only execution at higher concurrency.
Second, local inference increases task completion time due to longer model inference, while higher concurrency brings limited goodput gains when task success declines.
Third, local-only execution eliminates cloud model API costs, but hybrid execution does not necessarily cost less than cloud-only execution.
Fourth, local-only execution avoids exposing sensitive task information to cloud agents, while exposure in hybrid architectures depends on when cloud agents are involved and what information they receive.
Overall, there is no one-size-fits-all deployment for agentic AI on personal devices. 
They also motivate evaluating local execution on intended tasks, limiting concurrency when reliability matters, and exploring hybrid designs in which local agents lead and request targeted cloud assistance.

We make two main \textbf{contributions}: 

\textbf{(1) Performance characterization and empirical insights.} We systematically characterize the trade-offs of agent execution on personal AI devices and derive practical implications for deployment and performance optimization. 

\textbf{(2) Benchmark and open artifacts.} We introduce \name, a benchmark suite for evaluating agentic workloads on personal AI devices, and release our implementation and execution traces to enable reproducible evaluation and research in this nascent area.

The rest of this paper is organized as follows. 
Section~\ref{sec:related_work} considers related work. 
Section~\ref{sec:design} presents \name benchmark.
Section~\ref{sec:evaluation} highlights the results obtained from running \name across deployment architectures and concurrency levels. 
Section~\ref{sec:designimplications} considers the design implications.
Section~\ref{sec:dataset} presents the \name dataset release. 
Section~\ref{sec:conclusion} concludes this paper.

\section{Background and Related Work}
\label{sec:related_work}
\textbf{Agent Execution on Personal AI Devices.}
Advances in small language models and on-device AI accelerators have made local agent execution increasingly practical. Recent work have explored on-device agents. Agent-X~\cite{chung2026agentx} optimizes the end-to-end execution of on-device agents, while PalmClaw~\cite{cai2026palmclaw} runs the agent loop, memory, and tool use natively on mobile devices. These systems demonstrate the feasibility of local agent execution, but also identify performance limitations on resource-constrained devices.

A complementary line of work explores hybrid device--cloud execution. EcoAgent~\cite{yi2026ecoagent} uses cloud models for planning and executes actions on the device, whereas Hera~\cite{zhang2026hera} dynamically selects between device and cloud agents for individual steps of a task. Recent work has further considered how agent execution can be partitioned across device and cloud models to balance task success, performance, and cloud usage of agent systems~\cite{rainone2026cloudagents}.

Together, these studies show local and hybrid agent execution are viable alternatives to cloud-native agents. However, their benefits and limitations remain unclear.

\textbf{Agent Benchmarks.}
Existing benchmarks that evaluate agents fall broadly into two categories: \emph{capability-oriented} benchmarks that measure whether agents can complete realistic tasks, and \emph{systems-oriented} benchmarks that characterize the performance and resource costs of agent execution.

Capability-oriented benchmarks cover diverse and realistic workloads. For example, GAIA~\cite{mialon2024gaia} evaluates reasoning, multimodal, browsing, and tool-use tasks; TUA-Bench~\cite{chen2026tua} spans productivity and specialized professional workflows; Terminal-Bench~\cite{merrill2026terminal} focuses on technical tasks through command-line interface; MyPCBench~\cite{jang2026mypcbench} evaluates agents in personalized desktop environments. These benchmarks primarily measure task success.

Systems-oriented benchmarks in contrast examine how agents execute. For example, XPerf~\cite{wang2026XPerf} evaluates LLM serving performance using agent execution traces, while OSWorld-Human~\cite{abhyankar2026osworld} and AgentSysBench~\cite{chang2026agentsysbench} characterize end-to-end latency, component-level costs, and execution bottlenecks. However, neither line of work systematically explores the trade-offs that arise when agent execution moves from the cloud to personal AI devices.
This gap motivates \name, \textit{which combines task evaluation and
execution measurements within a common benchmark framework}.

\section{\name}
\label{sec:design}
This section presents the methodology of \name for characterizing agent execution on personal AI devices.We begin with an overview of the workflow, then describe the agent workloads, execution configurations, and how agents complete tasks under these configurations. We next present the measurements and evaluation metrics, followed by the experimental procedure used in this study.

\begin{figure*}[htp]
  \centering
  \includegraphics[width=\linewidth]{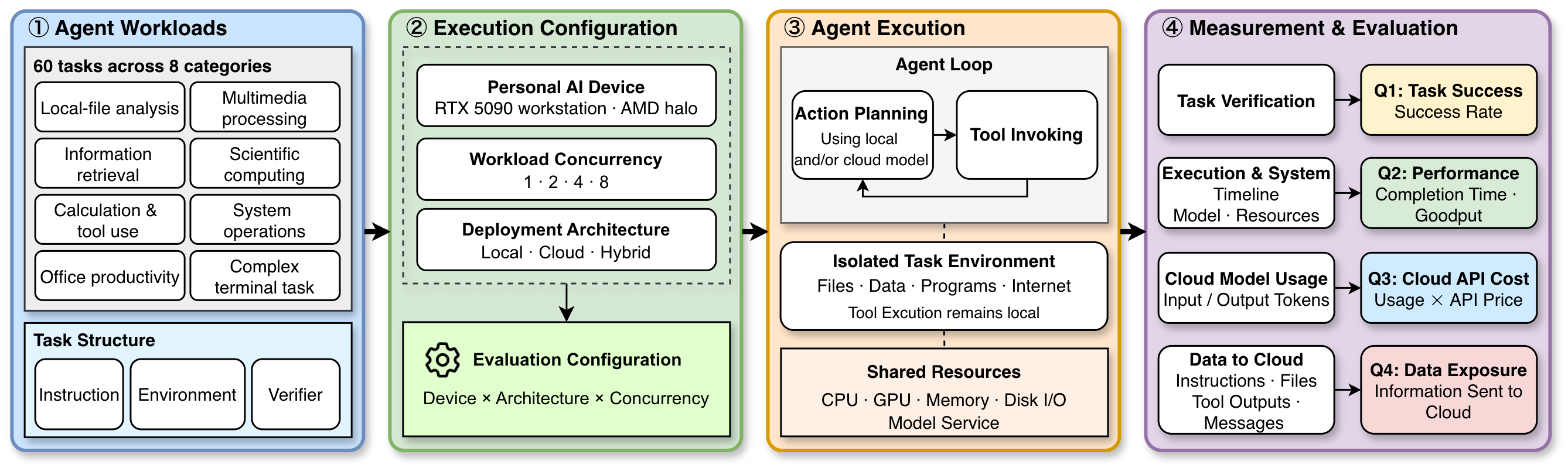}
  \caption{
    Overview of \name. Agent workloads are executed under controlled combinations of personal AI devices, deployment architectures, and workload concurrency levels. \name captures task outcomes, execution and system behavior, cloud model usage, and data sent to cloud for characterizing task success, execution performance, cloud API cost, and data exposure.
  }
  \label{fig:overview}
\end{figure*}

\subsection{Benchmark Overview}
\label{sec:benchmark-overview}
\name evaluates agent systems using a suite of 60 tasks across eight categories under specified deployment architectures, and workload concurrency configurations.
Figure~\ref{fig:overview} illustrates the workflow, which consists of four steps.

\textbf{Step 1.}
\name constructs a suite of agent tasks spanning diverse user activities and system demands, from information retrieval, file analysis, and office productivity to multimedia processing, scientific computing, and complex terminal operations. These tasks require different combinations of computation, memory, storage I/O, and network access. Each task provides a human-readable instruction, an isolated execution environment with the required resources, and a task-specific verifier.

\textbf{Step 2.}
Tasks are evaluated under controlled configurations.
Each configuration specifies a device, a deployment architecture, and a concurrency level.
The hardware resources of the devices are used for local model inference and tool execution, while the architecture determines where model inference occurs and how agents coordinate. The concurrency level sets the maximum number of tasks that can execute at the same time.


\textbf{Step 3.}
For each configuration, agents repeatedly invoke tools, and inspect the results until the task completes or terminates. Tool use includes file operations, program execution, information retrieval, and data processing.
Tasks run in isolated environments while sharing the underlying CPU, GPU, memory, storage, and, where applicable, the local model service, which enables characterizing end-to-end agent behavior and resource contention.

\textbf{Step 4.}
During execution, \name captures task outcomes, model and tool interactions, inter-agent communication, cloud model usage and data transfer, and system resource utilization. Task-specific verifiers assess returned answers and changes to the task environment. Together, these measurements characterize task success, end-to-end performance, cloud API cost, and data exposure, while fine-grained traces help explain differences across workloads and execution configurations.

\subsection{Agent Workloads}
\label{sec:agent-workloads}

\textbf{Workload coverage.}
We construct the workload suite to cover diverse user activities and system demands, ranging from information retrieval and office productivity to multimedia processing, scientific computing, and complex terminal tasks.
These workloads require different amounts of computation, memory,
storage I/O, and network access.
We select tasks from existing benchmarks~\cite{mialon2024gaia, chen2026tua, merrill2026terminal} that have fixed inputs, reproducible execution environments, and programmatically verifiable outcomes, excluding those that require human interaction or subjective assessment.
Among eligible tasks, we select a suite covering diverse personal computing activities, task difficulties, and resource demands.

\textbf{Task suite.}
The resulting suite contains 60 tasks drawn from GAIA~\cite{mialon2024gaia}, TUA-Bench~\cite{chen2026tua}, and Terminal-Bench 2~\cite{merrill2026terminal}, spanning eight task
categories. Table~\ref{tab:task-suite} summarizes the composition and characteristics of the suite. We adapt the selected tasks to \name's task format while preserving their original objectives. The complete task list is provided in Supplementary Section~A.


\begin{table}[t]
\centering
\caption{\name task categories and characteristics.}
\label{tab:task-suite}
\setlength{\tabcolsep}{2pt}
\footnotesize
\begin{tabularx}{\columnwidth}{
    @{}>{\raggedright\arraybackslash}p{2cm}
    >{\raggedright\arraybackslash}X
    >{\raggedright\arraybackslash}p{1.5cm}
    c@{}
}
\hline
\textbf{Task Category} & \textbf{Description} & \textbf{Requirements} & \textbf{\#Tasks} \\
\hline

Local-file analysis
& Analyze and combine local documents, spreadsheets, and archives.
& Memory, storage I/O
& 12 \\ \hline

Information retrieval
& Retrieve and analyze information from external sources.
& Network, computation
& 10 \\ \hline

Calculation and tool use
& Perform calculations and execute task-specific tools and programs.
& Computation, memory
& 8 \\ \hline

Office productivity
& Create, edit, and process documents, spreadsheets, and presentations.
& Memory, storage I/O
& 8 \\ \hline

Multimedia processing
& Process and transform images, audio, and video.
& Computation, memory, storage I/O
& 4 \\ \hline

Scientific and engineering computing
& Analyze scientific data and perform numerical computation.
& Computation, memory, storage I/O
& 4 \\ \hline

System and software operations
& Configure software environments and perform system-level operations.
& Computation, storage I/O
& 2 \\ \hline

Complex terminal problem solving
& Build, debug, and recover software and system artifacts through
multi-step operations.
& Computation, memory, storage I/O
& 12 \\

\hline
\textbf{Total} & & & \textbf{60} \\
\hline
\end{tabularx}
\end{table}

\textbf{Task structure.}
Each \name task comprises a human-readable instruction specifying the task goal and requirements for the agent, an isolated
execution environment, and a task-specific verifier. The environment provides the files, data, programs, and network access required to complete the task. The verifier assesses the final answer, artifact, or environment state against task-specific success criteria, which is inaccessible to the agent during execution.

\subsection{Execution Configuration}
\label{sec:execution-configuration}

\textbf{Personal AI devices.}
We evaluate \name on two personal AI devices. The first is a high-end workstation with an Intel Core Ultra 9 285K CPU and an NVIDIA GeForce RTX 5090 GPU, and the second is a compact AI system with an AMD Ryzen AI Max+ 395 processor, integrated Radeon 8060S GPU. As summarized in Table~\ref{tab:devices}, the two devices represent discrete- and integrated-GPU designs with dedicated and unified memory, respectively.

\begin{table}[t]
\centering
\caption{Specification of the evaluated personal AI devices.}
\label{tab:devices}
\footnotesize
\setlength{\tabcolsep}{4pt}
\renewcommand{\tabularxcolumn}[1]{m{#1}}

\begin{tabularx}{\columnwidth}{
    @{}l
    >{\raggedright\arraybackslash}X
    >{\raggedright\arraybackslash}X@{}
}
\hline
\textbf{Hardware}
& \textbf{RTX 5090}
& \textbf{Max+ 395} \\
\hline

Processor
& Intel Core Ultra 9 285K
& AMD Ryzen AI Max+ 395 \\

CPU cores / threads
& 24 / 24
& 16 / 32 \\

GPU
& NVIDIA GeForce RTX 5090
& AMD Radeon 8060S \\

GPU integration
& Discrete
& Integrated \\

System memory
& 128\,GB
& 128\,GB unified \\

GPU memory
& 32\,GB dedicated
& 96\,GB allocated \\
\hline
\end{tabularx}

\smallskip
\begin{minipage}{\columnwidth}
\footnotesize
The Max+ 395 allocates 96\,GB of unified memory to the GPU,
leaving approximately 30.5\,GB visible to the operating system in our
configuration.
\end{minipage}
\end{table}

\textbf{Concurrent workloads.}
We evaluate workload concurrency levels of $1$, $2$, $4$, and $8$, where
a concurrency level of $C$ allows up to $C$ tasks to execute simultaneously
on the same device. Varying concurrency allows us to characterize how
execution performance changes as workload intensity increases.

\begin{figure}[t]
  \centering
  \includegraphics[width=\linewidth]{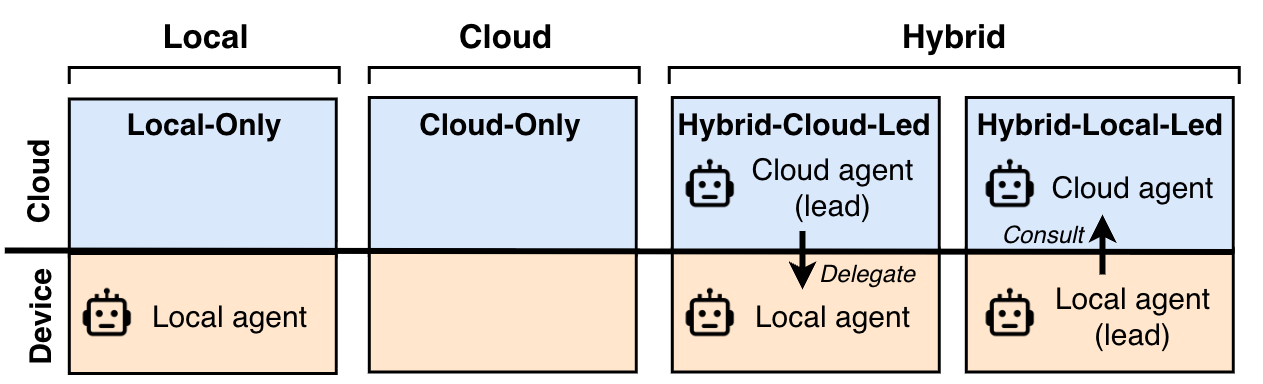}
  \caption{Deployment architectures evaluated by \name.}
  \label{fig:deployment-architectures}
  \Description{Deployment architectures evaluated by \name.}
\end{figure}

\textbf{Deployment architectures.}
We evaluate four deployment architectures that differ in where model
inference occurs and how local and cloud agents coordinate, as shown in
Figure~\ref{fig:deployment-architectures}. Across all architectures, task
environments and tool execution remain on the personal AI device.
\textbf{\textit{Local-Only (LO)}} uses a local agent for both action planning and tool invocation, here the model inference is performed locally.
\textbf{\textit{Cloud-Only (CO)}} uses a cloud agent for action planning, with model
inference performed through the cloud API while tool execution remains local.
\textbf{\textit{Hybrid Cloud-Led (HCL)}} follows a delegation-based design, where a cloud
agent leads task execution and decides when to delegate actions to a local
agent. The local agent uses local model inference to complete delegated
actions and returns the results to the cloud agent.
\textbf{\textit{Hybrid Local-Led (HLL)}} follows a consultation-based design, where a local agent leads task execution and consults a cloud agent for reasoning when needed.
The cloud agent returns its response to the local agent, which
continues execution locally.  Across all architectures, task environments and tool execution remain on the personal AI device.

\textbf{Inference setup.}
For local inference, we run Qwen3.8-27B~\cite{qwen38} with UD-Q6\_K\_L 6-bit quantization~\cite{unsloth_qwen38_gguf} using llama.cpp (commit 0b5be7e4)~\cite{llamacpp}, with CUDA 12.8.1 on the RTX 5090 and Vulkan on the Max+ 395. We enable Flash Attention~\cite{dao2022flashattention}, an 8-bit KV cache, and multi-token prediction (MTP)~\cite{gloeckle2024mtp} speculative decoding with up to two draft tokens, and allow up to eight concurrent model calls on each device. For cloud inference, we use DeepSeek V4 Flash~\cite{deepseekv4} with high reasoning effort. The local and cloud models use context windows of 65,536 and 1,000,000 tokens, with maximum outputs of 8,192 and 384,000 tokens per call, respectively. These settings are fixed throughout the evaluation. Further configuration details are provided in Supplementary Section~C, and the unit prices used to estimate cloud API costs in Supplementary Section~D.

The experiments comprise 32 configurations, covering all combinations of two personal AI devices, four deployment architectures, and four task concurrency levels.
Executing the complete 60-task suite once under each configuration produces 1,920 task executions, generating approximately $162.07$ million raw execution and measurement records.

\subsection{Agent Execution}
\label{sec:agent-execution}

\textbf{Agent loop.}
Each task follows an iterative loop in which the agent uses its model to determine the next action, invokes tools when needed, and uses the returned results to continue execution until the task completes or terminates. Hybrid architectures additionally allow agents to delegate or consult through the same loop. We implement the agent loop using Pi Coding Agent v0.84.1~\cite{zechner2026pi}.

\textbf{Task environment.}
Each task runs in an isolated Docker container with its required files, data, programs, and dependencies. Agents interact with the environment via a common tool interface for file operations, command execution, information retrieval, and data processing. Tool access is defined by the agents' system prompts, which remain fixed throughout the evaluation. The system prompts are provided in Supplementary Section C. The environment is reset before each execution.

\textbf{Shared resources.}
Local model inference is provided by shared model serving, while cloud inference is accessed through the cloud API. Concurrent tasks remain isolated at the environment level but share CPU, GPU, memory, storage I/O,
and the local model service.

\subsection{Measurement}
\label{sec:measurement}

\textbf{Primary metrics.}
We evaluate agent execution along four dimensions corresponding to our research questions: task success, execution performance, cloud API cost, and cloud data exposure. Table~\ref{tab:measurements} summarizes the primary metrics used in our evaluation.

Task success is determined by task-specific verifiers, while execution performance is characterized by mean completion time on tasks successfully completed by all architectures within each device--concurrency setting, and by goodput.
For cloud API cost, we additionally distinguish costs incurred by successful and unsuccessful executions. 
To assess cloud data exposure, we examine task instructions, task-file contents, tool outputs, and inter-agent messages sent to cloud models. Information processed by local models or tools is excluded.
We first manually identify sensitive items in the initial task inputs.
Each item represents a distinct piece of sensitive information, such as a private email or an authentication credential.
We report cloud data exposure as the number of distinct sensitive items observed in recorded cloud inputs divided by the total number of sensitive items identified in the initial task inputs..

\textbf{Diagnostic measurements.}
\name collects fine-grained measurements to explain performance differences as summarized in Table~\ref{tab:measurements}. They capture model and tool execution times, inter-agent interactions, model usage and serving
performance, and system resource utilization. We use these measurements to
quantify execution time and identify performance bottlenecks across
deployment architectures and concurrency levels. 
Supplementary Section~B details the execution, model usage, and resource records.

\begin{table}[t]
\centering
\caption{Evaluation metrics and diagnostic measurements.}
\label{tab:measurements}
\setlength{\tabcolsep}{3pt}
\footnotesize

\begin{tabularx}{\columnwidth}{
    @{}c|p{1.72cm}|X@{}
}
\hline
\textbf{Type} & \textbf{Metric} & \textbf{Measurement} \\
\hline

\multirow{5}{*}{\rotatebox[origin=c]{90}{Primary}}
& Task Success
& Fraction of tasks passing their task-specific verifiers. \\

& Completion Time
& Wall-clock time from task start to termination. \\

& Goodput
& Successfully completed tasks per unit of benchmark time. \\

& API Cost
& Cloud model charges based on usage and provider pricing. \\

& Data Exposure
& Fraction of sensitive information exposed to cloud agents. \\

\midrule

\multirow{3}{*}{\rotatebox[origin=c]{90}{Diagnostic}}
& Execution
& Model and tool execution times, inter-agent interactions, and timestamps. \\

& Model
& Input/output tokens, prefill/decode time, and throughput. \\

& System
& CPU/GPU utilization, memory usage, and storage I/O. \\

\hline
\end{tabularx}
\end{table}

\subsection{Evaluation Method}
\label{sec:evaluation_procedure}

We run all 60 tasks under each of the 32 configurations defined in Section~3.3, resulting in 1,920 task runs in total. Tasks are executed in the same order across configurations using a fixed random seed. We do not limit the number of agent-loop iterations, tool calls, delegations, or consultations. Instead, each task has a 7,200-second time limit shared by all participating agents.

Our main analysis is based on one complete run of the full configuration matrix, requiring approximately 440 device hours. We further assess the stability of the results with two additional runs of the complete 60-task suite for each architecture on the RTX 5090 at \(C=1\) and \(C=8\). Across the three runs, success counts vary by at most 6 out of 60 tasks, while the overall trends in task success and goodput remain consistent. Detailed results are provided in Supplementary Section E.

\section{Results and Discussion}
\label{sec:evaluation}
This section addresses Q1--Q4 by examining task success, execution performance, cloud API cost, and cloud data exposure across deployment architectures, devices, and concurrency levels. We then examine the trade-offs among these outcomes.
Additional results and execution measurements are in Supplementary Section~F.

\subsection{Task Success}
\label{sec:task-success}

\begin{figure}[t]
  \centering
  \includegraphics[width=0.9\linewidth]{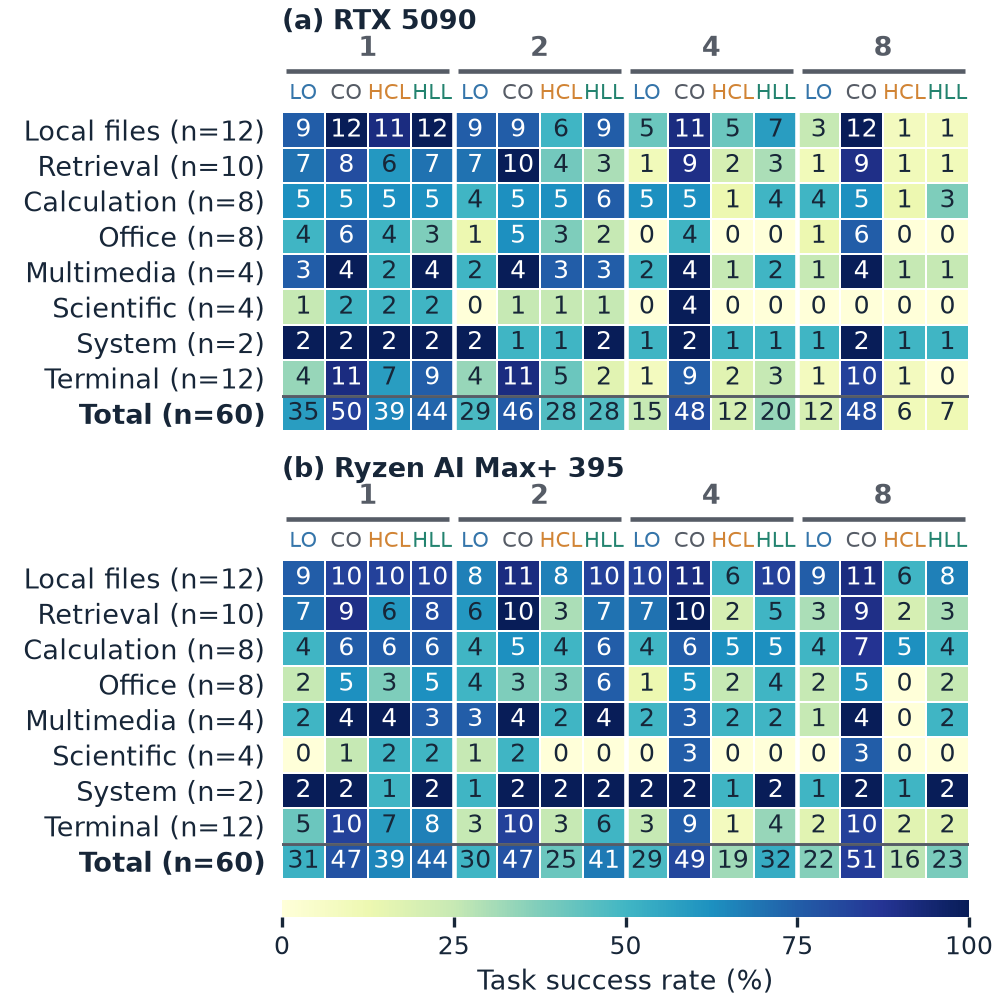}
  \caption{Task success by workload category.}
  \label{fig:task-success}
    \Description{Task success by workload category.}
\end{figure}

\textbf{Overall Task Success.}
Figure~\ref{fig:task-success} compares the number of successful tasks across the four deployment architectures at different concurrency levels.
Overall, LO completes more than half of the 60 tasks at low concurrency, but its success drops sharply as concurrency increases, whereas CO remains comparatively stable.
On the RTX 5090, LO drops fro m 35 successful tasks at concurrency 1 to 12 at concurrency 8, while CO changes only from 50 to 48.
On the Max+ 395, LO drops from 31 to 22, while CO increases from 47 to 51.
At concurrency 1, both hybrid architectures improve on LO: HCL and HLL complete 39 and 44 tasks on each device, respectively.
Their advantage is inconsistent at higher concurrency. At concurrency 8, both hybrids fall below LO on the RTX 5090, while HCL falls below LO and HLL remains only marginally above it on the Max+ 395.

\textbf{Success across Tasks.}
Aggregate success counts conceal differences across task categories.
Figure~\ref{fig:task-success} breaks down successful tasks by category for each architecture and concurrency level.
At $C=1$, LO completes 9 of 12 local-file tasks and 7 of 10 retrieval tasks on both systems.
On the RTX 5090, it also matches CO in calculation (5/8) and system tasks (2/2).
The largest gap appears in terminal tasks: LO completes 4 of 12 on the RTX 5090 and 5 of 12 on the Max+ 395, compared with 11 and 10 for CO.
HCL and HLL narrow this gap, completing 7 and 9 terminal tasks on the RTX 5090, and 7 and 8 on the Max+ 395, respectively.
Thus, the overall success gap varies considerably with the task mix.

\begin{figure}[t]
  \centering
  \includegraphics[width=\linewidth]{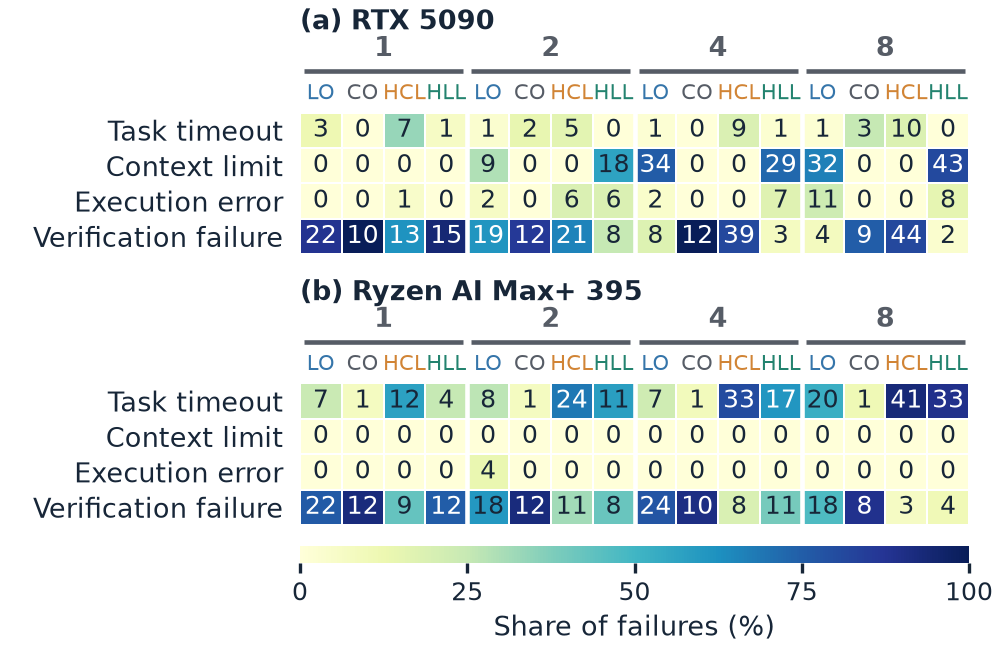}
  \caption{Failure analysis.}
  \label{fig:failure-analysis}
  \Description{Failure analysis.}
\end{figure}

\textbf{Failure Analysis.}
Figure~\ref{fig:failure-analysis} breaks down unsuccessful tasks by failure type across devices, architectures, and concurrency levels.
At $C=8$ on the RTX 5090, LO records 32 context-limit errors and 11 execution errors, while HLL records 43 context-limit errors.
Concurrent requests compete for a shared KV-cache pool, so pool exhaustion can cause a context error even when a request remains below its own sequence limit.
On the Max+ 395, no context-limit errors are recorded; instead, slower local inference contributes to timeouts for LO, HCL, and HLL (20, 41, and 33 tasks, respectively).

The two hybrid architectures differ in how local context errors appear in the results.
At $C=8$ on the RTX 5090, HCL records 44 verification failures but no context-limit errors.
In all 44 cases, its local component encounters a context error, but the cloud agent subsequently returns a final response that fails verification.
Thus, HCL's zero recorded context-limit errors do not indicate that its local component avoids context exhaustion.
CO records no context-limit errors, and most of its failures are verification failures.

Local context errors can also lead to timeouts in HCL.
The local agent does not automatically handle some context errors from the local model service, such as by compacting its message history, and retains its prior history.
The cloud agent typically receives a generic execution-failure report and may continue delegating work to the same local agent until the task deadline.
For example, at $C=8$ on RTX 5090, ten HCL tasks encounter repeated context errors, yet the cloud agent issues over 100 delegations per task without resolving them.
All ten time out at the two-hour deadline.

\begin{boxH}
\textbf{Observation on Q1:} 
At low concurrency, LO achieves lower overall task success than CO, but performs comparably on particular task categories.
The gap becomes significant as concurrency increases, driven by context errors and timeouts.
Hybrid architectures narrow the gap at low concurrency, but their gains do not consistently persist at higher concurrency.
\end{boxH}

\subsection{Execution Performance}
\label{sec:results-performance}

\textbf{Execution Time.}
Table~\ref{tab:mean-successful-completion-time} reports mean completion time for tasks successfully completed by all four architectures.
On both devices, CO completes these tasks substantially faster than LO and the two hybrid architectures.
At $C=1$ on the RTX 5090, CO completes these tasks in 2.07 minutes, compared to 7.01 minutes for LO, 6.75 minutes for HLL, and 15.80 minutes for HCL.
On Max+ 395, the corresponding times are 1.95, 15.15, 17.28, and 34.14 minutes.

Notably, HCL takes longer than LO on both devices. In HCL, the cloud agent delegates additional work, including extensive preliminary checks, to the local agent and waits for it to finish. For example, on Max+ 395, the local agent spends 93.5 minutes on preliminary checks for an MP3 metadata-editing task before making any changes; the task then times out. HLL, by contrast, has a mean completion time closer to LO on the RTX 5090. Cloud assistance therefore does not necessarily shorten completion time when execution still depends on the local agent.

\begin{table}[t]
  \centering
  \caption{Mean completion time (min) on tasks completed successfully by all four architectures within each device--concurrency setting. All four entries in a column use the same $n$ tasks. }
  \label{tab:mean-successful-completion-time}
  \footnotesize
  \begin{tabular*}{\columnwidth}{@{\extracolsep{\fill}}lcccccccc@{}}
    \hline
    \multirow{3}{*}{\textbf{Architecture}} & 
    \multicolumn{8}{c}{\textbf{Task concurrency}}
    \\
    \cline{2-9}
    & \multicolumn{4}{c}{\textbf{RTX 5090}} & \multicolumn{4}{c}{\textbf{Max+ 395}} \\
    \cline{2-5}
    \cline{6-9}
    & \textbf{1} & \textbf{2} & \textbf{4} & \textbf{8} & \textbf{1} & \textbf{2} & \textbf{4} & \textbf{8} \\
    \hline
    $n$ & 29 & 12 & 6 & 3 & 21 & 18 & 14 & 14 \\
    \hline
    LO & 7.01 & 3.37 & 4.88 & 3.88 & 15.15 & 12.81 & 16.87 & 13.67 \\
    CO & 2.07 & 0.72 & 2.00 & 0.18 & 1.95 & 1.27 & 0.75 & 0.41 \\
    HCL & 15.80 & 9.88 & 22.31 & 39.03 & 34.14 & 31.07 & 26.41 & 36.32 \\
    HLL & 6.75 & 3.30 & 6.32 & 3.47 & 17.28 & 10.99 & 12.02 & 23.86 \\
    \hline
  \end{tabular*}
\end{table}

\begin{table}[t]
  \centering
  \caption{Elapsed wall-clock time (h) for executing the full 60-task suite.}
  \label{tab:full-suite-run-time}
  \footnotesize
  \begin{tabular*}{\columnwidth}{@{\extracolsep{\fill}}lcccccccc@{}}
    \hline
    \multirow{3}{*}{\textbf{Architecture}} & 
    \multicolumn{8}{c}{\textbf{Task concurrency}}
    \\
    \cline{2-9}
    & \multicolumn{4}{c}{\textbf{RTX 5090}} & \multicolumn{4}{c}{\textbf{Max+ 395}} \\
    \cline{2-5}
    \cline{6-9}
    & \textbf{1} & \textbf{2} & \textbf{4} & \textbf{8} & \textbf{1} & \textbf{2} & \textbf{4} & \textbf{8} \\
    \hline
    LO & 16.66 & 6.03 & 3.49 & 2.62 & 31.43 & 17.93 & 12.59 & 8.91 \\
    CO & 4.12 & 4.99 & 1.99 & 2.49 & 8.56 & 2.50 & 2.68 & 2.48 \\
    HCL & 33.14 & 21.65 & 10.02 & 5.84 & 58.88 & 35.05 & 21.10 & 12.66 \\
    HLL & 14.83 & 4.96 & 2.76 & 1.00 & 36.70 & 23.36 & 16.33 & 11.82 \\
    \hline
  \end{tabular*}
\end{table}

\begin{figure}[t]
  \centering
  \includegraphics[width=\linewidth]{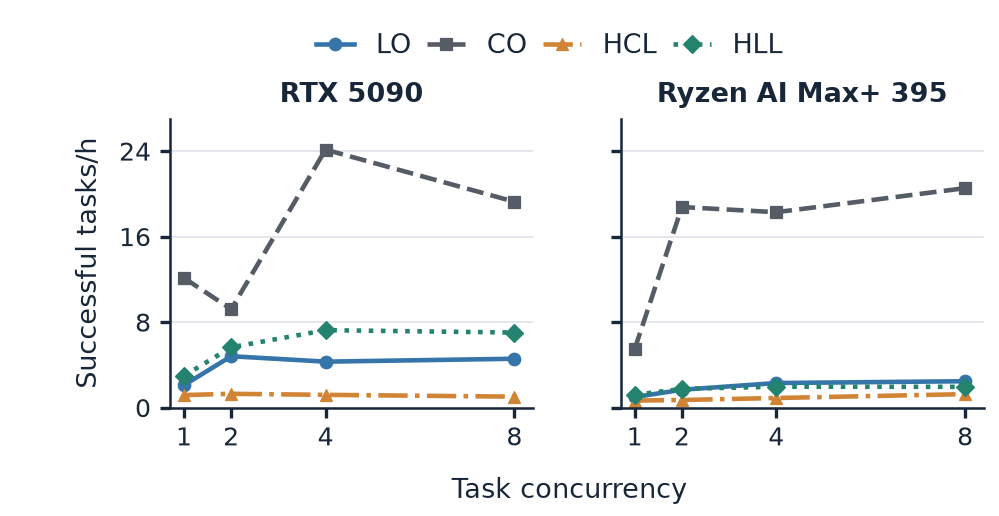}
  \caption{Goodput (no. of successfully completed tasks / total run time).}
  \label{fig:goodput-concurrency}
  \Description{Goodput (no. of successfully completed tasks / total run time).}
\end{figure}

\textbf{Goodput.}
Figure~\ref{fig:goodput-concurrency} shows goodput, measured as successful tasks per hour of suite execution.
Table~\ref{tab:full-suite-run-time} reports the elapsed wall-clock time for executing the complete 60-task suite.
From $C=1$ to $C=8$ on the RTX 5090, LO's suite execution time falls from 16.66 to 2.62 hours, a 6.35-fold reduction, but its goodput rises from 2.1 to 4.6 tasks/h, only a 2.18-fold increase.
For CO, suite execution time falls from 4.12 to 2.49 hours, a 1.65-fold reduction, while goodput rises from 12.1 to 19.3 tasks/h, a 1.59-fold increase.
On Max+ 395, LO and CO achieve respective execution-time reductions of 3.53-fold and 3.45-fold, while their goodput increases 2.50-fold and 3.74-fold.
Thus, although all architectures have shorter elapsed wall-clock times at $C=8$ than at $C=1$, only CO achieves roughly proportional goodput gains.
For architectures involving local agents, higher concurrency reduces the number of successfully completed tasks, limiting the benefit of shorter execution times.



\begin{figure}[t]
  \centering
  \includegraphics[width=0.9\linewidth]{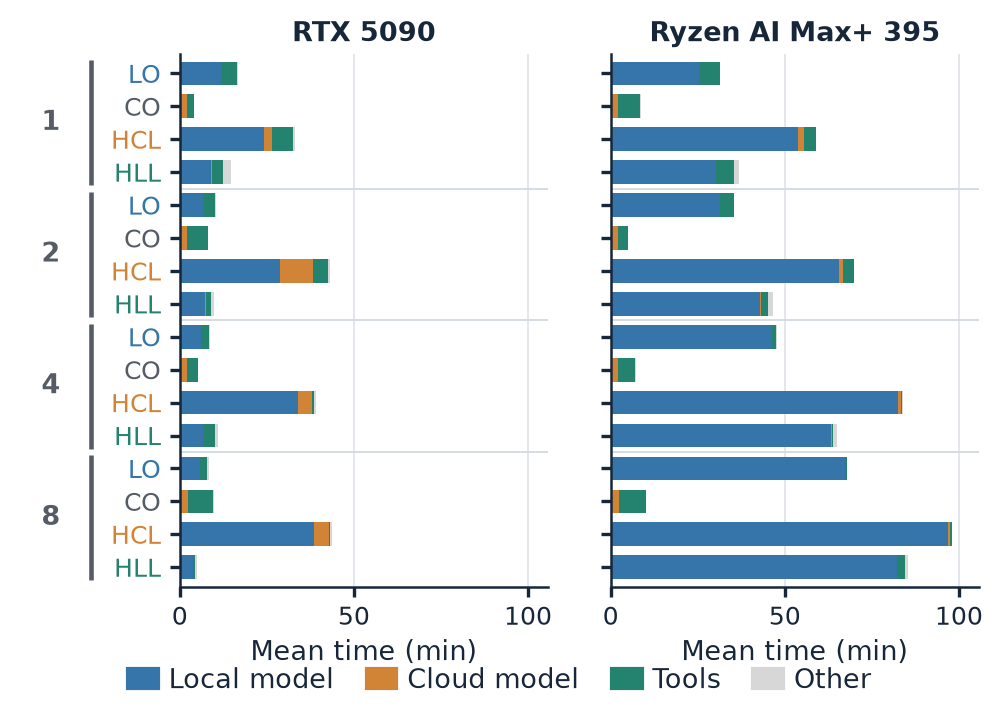}
  \caption{Breakdown of mean execution time per task.}
  \label{fig:execution-time-breakdown}
  \Description{Breakdown of mean execution time per task.}
\end{figure}

\textbf{Execution Time Breakdown.}
Figure~\ref{fig:execution-time-breakdown} shows the breakdown of mean task execution time.
LO, HCL, and HLL spend most of their task execution time waiting for responses from the local model.
For HCL on Max+ 395, the mean time spent waiting for local model responses per task increases from 53.9 minutes at $C=1$ to 96.9 minutes at $C=8$. Among HCL tasks that time out at $C=8$, waiting for local model responses accounts for approximately 99\% of execution time, leaving little time for tool execution.
In contrast, local tool execution can become the main source of delay for CO, particularly at higher concurrency. At $C=8$ on RTX 5090, CO spends an average of 7.4 minutes per task executing or waiting for tools, compared with 2.2 minutes waiting for cloud model responses.
Long-running tool calls can also become a bottleneck for task completion, consuming much of the execution budget without producing a successful outcome.
For example, at $C=8$ on RTX 5090, three CO tasks each spend more than 110 minutes executing or waiting for tools and ultimately time out at the two-hour deadline.

\begin{boxH}
\textbf{Observation on Q2:}
    Local inference is the central performance constraint: it makes LO much slower than CO, and cloud assistance does little to offset it—HLL takes about as long as LO on the RTX 5090, while HCL is slower on both devices. Increasing concurrency improves goodput, but declining task success limits the gains for architectures that rely on local agents. CO, whose runtime is dominated by tool execution, avoids this trade-off.
\end{boxH}

\subsection{Cloud API Cost}
\label{sec:results-cost}

\begin{figure}[t]
  \centering
  \includegraphics[width=0.9\linewidth]{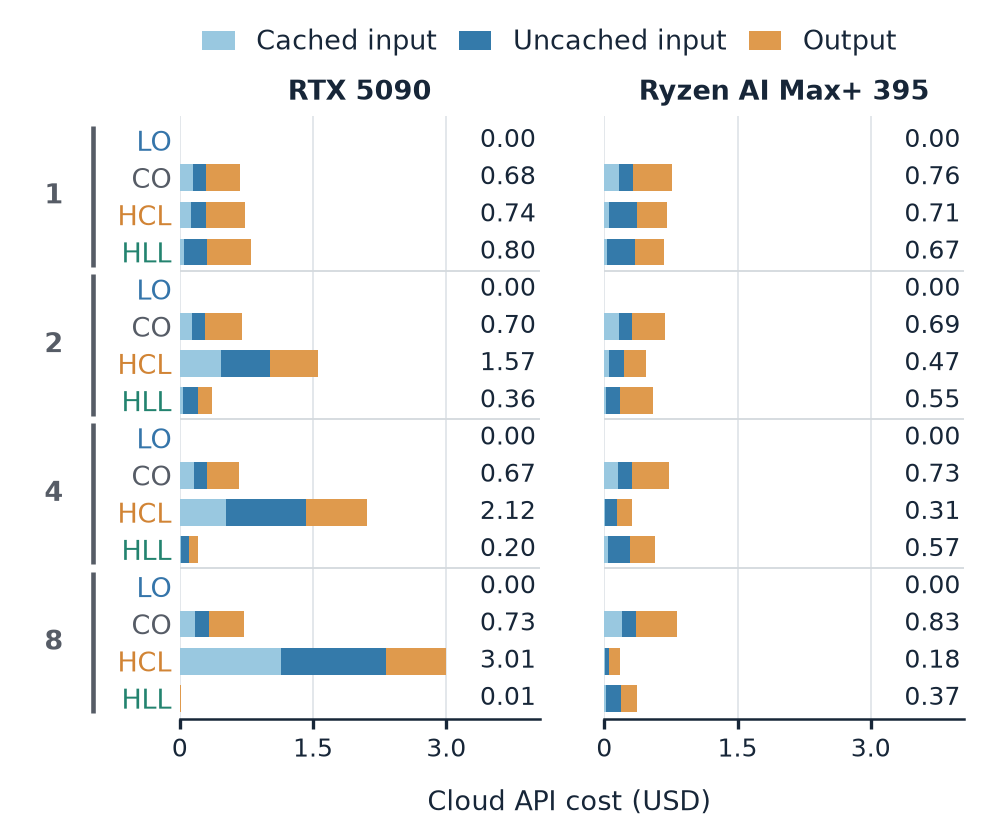}
  \caption{Cloud API cost breakdown.}
  \label{fig:cloud-cost-breakdown}
  \Description{Cloud API cost breakdown.}
\end{figure}

Figure~\ref{fig:cloud-cost-breakdown} breaks down cloud API costs into cached-input, uncached-input, and output-token charges. The hybrid architectures do not always cost less than CO, even though local agents perform part of the work. There are two reasons. First, cloud agents still need the task context and the results of local work to decide what to do next, so delegating work does not necessarily reduce cloud input tokens in proportion to the work delegated. Second, API costs depend on the types of tokens used: uncached input and generated output cost more than cached input.

At $C=1$ on the RTX 5090, for example, HLL uses fewer cloud input tokens than CO (Table~\ref{tab:overall-cloud-exposure}). Yet 8.8\% of HLL's input is uncached, compared with 2.1\% for CO, and HLL generates 30.2\% more output tokens. The higher uncached-input and output charges exceed its savings on cached input, making HLL more expensive overall. HCL highlights another source of cost: additional cloud calls during coordination. On the RTX 5090, its recorded cloud calls rise from 770 at $C=1$ to 3,206 at $C=8$. Input charges account for 89.5\% of the associated cost increase, suggesting that the extra calls add substantial input-processing costs.

\begin{table}[t]
  \centering
  \caption{Recorded cloud input tokens (in millions, rounded to the nearest integer), including repeated context.}
  \label{tab:overall-cloud-exposure}
  \footnotesize
  \begin{tabular*}{\columnwidth}{@{\extracolsep{\fill}}lcccccccc@{}}
    \hline
    \multirow{3}{*}{\textbf{Architecture}} & \multicolumn{8}{c}{\textbf{Task concurrency}} 
    \\
    \cline{2-9}
    & 
    \multicolumn{4}{c}{\textbf{RTX 5090}} & \multicolumn{4}{c}{\textbf{Max+ 395}}
    \\
    \cline{2-5}
    \cline{6-9}
    & \textbf{1} & \textbf{2} & \textbf{4} & \textbf{8} & \textbf{1} & \textbf{2} & \textbf{4} & \textbf{8} \\
    \hline
    LO & 0 & 0 & 0 & 0 & 0 & 0 & 0 & 0 \\
    CO & 52 & 52 & 57 & 63 & 60 & 59 & 58 & 72 \\
    HCL & 46 & 169 & 195 & 418 & 21 & 18 & 4 & 2 \\
    HLL & 21 & 13 & 6 & 0 & 15 & 10 & 15 & 10 \\
    \hline
  \end{tabular*}
\end{table}

\begin{boxH}
\textbf{Observation on Q3:} 
    LO eliminates cloud-model API charges.
    Hybrid execution does not guarantee lower cloud API cost, because cloud agents still need relevant context, while higher uncached-input and output charges, together with additional coordination and recovery attempts, can offset potential savings.
\end{boxH}

\subsection{Cloud Data Exposure}
\label{sec:results-exposure}
To measure cloud data exposure, we examine whether sensitive information in the task suite appears in the inputs sent to cloud agents. We first inspect the task instructions and provided files and identify 527 distinct sensitive items, including personal identifiers, credentials, personal records, and confidential business information. We then use GPT-5.6 Sol~\cite{openai_gpt56_sol} with high reasoning effort to locate these items in the recorded cloud-agent inputs and manually verify the matches. We count each sensitive item at most once per evaluation configuration, regardless of how often it appears in cloud-agent inputs. Supplementary Section~G describes the sensitivity definitions and calculation procedure.

Table~\ref{tab:sensitive-data-exposure} shows the percentage of identified sensitive items that appear in cloud-agent inputs under each evaluation configuration. LO exposes none of these items, while CO exposes 76\% across all configurations. At $C=1$, HLL exposes 54 and 63 percentage points fewer items than CO on the RTX 5090 and Max+ 395, respectively. HCL reduces exposure by only 5 and 9 percentage points. This difference reflects when cloud agents enter the workflow. In HCL, the cloud agent directs the task and receives task files and reports from the local agent. In HLL, the local agent works first and can complete some tasks without calling a cloud agent. From $C=1$ to $C=8$, HLL's exposure drops from 22\% to 0\% on the RTX 5090 and from 13\% to 2\% on Max+ 395, while HCL's exposure remains high. This decline may partly reflect lower task success: some tasks fail before sharing sensitive information with the cloud.

\begin{table}[t]
  \centering
  \caption{Observed sensitive-information exposure to cloud agents (\%). Percentages are calculated relative to the 527 sensitive items identified in the initial inputs of the task suite.}
  \label{tab:sensitive-data-exposure}
  \footnotesize
  \begin{tabular*}{\columnwidth}{@{\extracolsep{\fill}}lcccccccc@{}}
    \hline
    \multirow{3}{*}{\textbf{Architecture}} &
    \multicolumn{8}{c}{\textbf{Task concurrency}} \\
    \cline{2-9}
    & \multicolumn{4}{c}{\textbf{RTX 5090}}
    & \multicolumn{4}{c}{\textbf{Max+ 395}} \\
    \cline{2-5}\cline{6-9}
    & \textbf{1} & \textbf{2} & \textbf{4} & \textbf{8}
    & \textbf{1} & \textbf{2} & \textbf{4} & \textbf{8} \\
    \hline
    LO & 0 & 0 & 0 & 0 & 0 & 0 & 0 & 0 \\
    CO & 76 & 76 & 76 & 76 & 76 & 76 & 76 & 76 \\
    HCL & 71 & 63 & 70 & 17 & 67 & 76 & 66 & 65 \\
    HLL & 22 & 21 & 13 & 0 & 13 & 11 & 22 & 2 \\
    \hline
  \end{tabular*}
\end{table}


\begin{boxH}
\textbf{Observation on Q4:}
    Local agents reduce sensitive-information exposure most when they can complete tasks before involving the cloud. If a cloud agent directs the work, it may need access to task files and local execution results, limiting the reduction.
\end{boxH}

\subsection{Cross-Metric Trade-offs}
\label{sec:trade-offs}
\begin{figure}[t]
  \centering
  \includegraphics[width=0.9\linewidth]{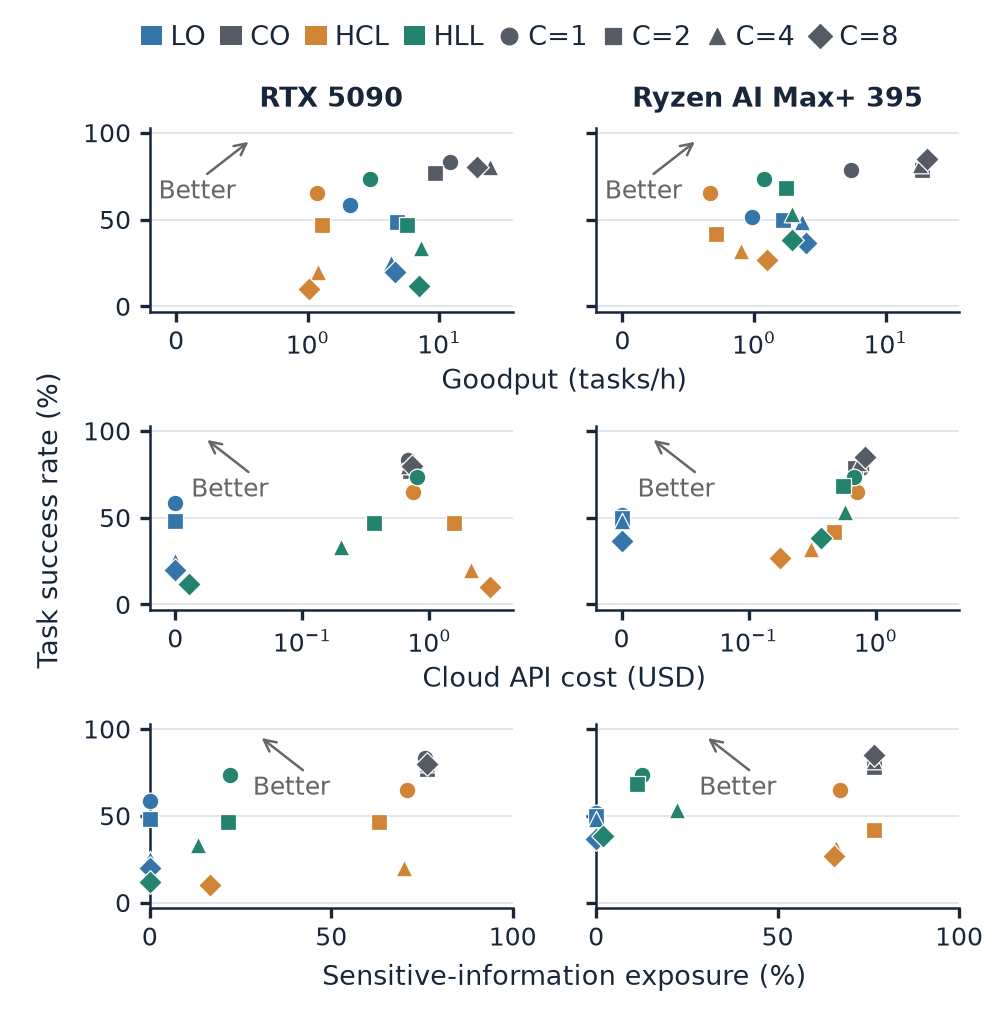}
  \caption{Cross metric tradeoffs comparing task success rate against goodput, cloud API cost and  sensitive-information exposure.}
  \label{fig:cross-metric-tradeoffs}
  \Description{Cross metric tradeoffs comparing task success rate against goodput, cloud API cost and  sensitive-information exposure.}
\end{figure}

Figure~\ref{fig:cross-metric-tradeoffs} compares task success rate with goodput, cloud API cost, and sensitive-information exposure across evaluation configurations. CO achieves higher task success and goodput than LO, while LO incurs no cloud model API cost and exposes no sensitive items to cloud agents. HLL can recover some of LO's lost task success while keeping cloud cost and exposure below CO's, but it does not match CO's goodput. At $C=1$ on Max+ 395, for example, HLL approaches CO's task success rate and reduces both cloud API cost and exposure, yet completes far fewer successful tasks per hour. Increasing concurrency can raise goodput, but for architectures that use local agents, it can also reduce task success. Thus, CO is preferable when completing tasks quickly and reliably matters most, whereas LO or HLL may be preferable when limiting cloud cost and sensitive-information exposure matters more.

\begin{boxH}
\textbf{Overall Takeaway:}
Moving agents from the cloud to personal AI devices eliminates cloud API costs and sensitive-information exposure in local-only execution, but reduces task success and lengthens completion time. Cloud assistance can recover some task success and reduce exposure relative to cloud-only execution, yet it offers little improvement in completion time and does not close the goodput gap, especially at higher concurrency.
\end{boxH}

\section{Design Implications}
\label{sec:designimplications}
The results point to three practical considerations for deploying agents on personal AI devices.

\textbf{Local Deployment Should Be Evaluated on the Intended Tasks.}
LO has lower overall task success and goodput than CO, but the success gap varies across task categories. At low concurrency, LO approaches CO's success rate in some categories and completes some tasks that CO fails. Aggregate results therefore cannot tell whether local execution will work well for a particular application. Before deployment, users should test local execution on the tasks they expect to run and check whether it meets their requirements for task success and completion time. If it does, they can avoid cloud model API costs and keep task information off cloud models.

\textbf{Concurrency Should Be Limited on Personal AI Devices.}
On the devices we evaluated, increasing concurrency improves goodput for LO and the hybrids, but it also reduces task success, particularly at higher concurrency levels. For applications that prioritize reliable completion, running one agent task at a time is therefore a sensible default on personal AI devices. Higher concurrency should be used only after testing whether its goodput gains justify the drop in task success on the target device and workload.

\textbf{Local-Led Hybrid Execution May Offer a Better Balance.}
At low concurrency, both hybrid architectures complete more tasks than LO. HLL has similar or slightly higher task success than HCL, completes tasks faster, and exposes much less sensitive information. The key difference is who leads the task: in HLL, the local agent works first and asks the cloud agent for help when needed; in HCL, the cloud agent directs the local agent and receives task files and progress reports. Hybrid systems may therefore benefit from letting the local agent lead and sending only the information needed when it asks for help. Future systems could request cloud help when local progress stalls, errors recur, or resources become constrained, sending only the information needed for that step. They should then assess whether this improves task success without substantially increasing completion time, cloud API cost, or data exposure.

\section{Benchmark Artifacts and Execution Traces}
\label{sec:dataset}
We release the benchmark artifacts and execution traces to support
reproducible evaluation and further research on agent systems running
on personal AI devices.

\textbf{Released Artifacts.}
We release the \name task suite, execution framework, traces, and system measurements at \url{https://anonymous.4open.science/r/AgBench-2777/}.
The task suite provides task instructions, environment configurations, and task-specific verifiers.
The framework implements the four agent architectures and includes model-serving configurations and code for collecting traces, model usage statistics, and resource measurements.

\textbf{Execution Trace Dataset.}
The execution trace dataset contains records from evaluations of four agent architectures on two personal AI devices at concurrency levels of 1, 2, 4, and 8.
It records model messages, tool calls and results, inter-agent communication, timestamps and model usage, capturing each task's execution sequence.
Records are organized by run and task execution.
Run configurations identify the device, agent architecture, and concurrency level, while task-execution and agent identifiers associate recorded interactions with the corresponding tasks and agents.
Verification results and execution measurements are included.
Timestamped system and local model serving measurements record device- and container-level resource usage and can be aligned with the execution traces.
The released dataset totals approximately 115 GB uncompressed and is distributed as 9.0 GB of compressed archives.
An accompanying README documents the record fields, measurement units, and identifiers used to link the records.

\section{Conclusion}
\label{sec:conclusion}
This paper presents \name for characterizing agent execution on personal AI devices using a suite of 60 tasks covering eight categories. We consider four deployment architectures, namely the local-only, cloud-only, hybrid cloud-led, and hybrid local-led, on two personal AI devices with different hardware accelerators under different concurrency levels. 
The devices can complete a large subset of the tasks locally, but broader task coverage and effective concurrent execution remain challenging.
At low concurrency, local execution has near similar success as on the cloud in some categories, but the overall success gap widens at higher concurrency.
Local inference generally increases execution time, and higher goodput does not always preserve task success.
Fully local execution avoids cloud API charges and sensitive information transfer to the cloud.
Hybrid execution can improve task success, but does not guarantee lower API cost relative to cloud-only inference.

These findings suggest assessing readiness of personal AI devices against the requirements of the intended usage scenario.
Local execution may suffice where it meets required task success and completion time, while other scenarios may benefit from cloud assistance.
Achieving this balance requires deployment choices informed by task requirements, resource-aware concurrency management, and hybrid coordination that adapts work allocation and information exchange to execution progress.






\balance
\bibliographystyle{ACM-Reference-Format} 
\bibliography{references}

\clearpage
\makeatletter
\@ACM@balancefalse
\makeatother
\nobalance

\appendix
\onecolumn
\section*{Supplementary Material}
\section{\name Task List}
\label{app:task-list}


\begin{table}[h]
\centering
\caption{The 60 tasks in \name and their resource profiles.}
\label{tab:effbench-task-list}
\footnotesize
\setlength{\tabcolsep}{3pt}
\begin{tabular*}{\textwidth}{@{\extracolsep{\fill}}
    >{\raggedright\arraybackslash}p{0.19\textwidth}
    >{\raggedright\arraybackslash}p{0.64\textwidth}
    >{\raggedright\arraybackslash}p{\dimexpr0.17\textwidth-4\tabcolsep\relax}@{}}
\hline
\textbf{Task Category} & \textbf{Task Summary} & \textbf{Resource Profile} \\
\hline
\multicolumn{3}{l}{\textbf{GAIA}} \\
\hline
\multirow[c]{12}{=}{Local-file analysis} & Extract the recommended reading page numbers from an MP3 recording. & Mixed \\
\cline{2-3}
 & Find the minimum number of cell towers needed to cover houses along a road. & Low footprint \\
\cline{2-3}
 & Determine whether colored spreadsheet cells admit a closed path without revisiting a cell. & Low footprint \\
\cline{2-3}
 & Follow an Excel map and identify the cell color reached on the specified turn. & Low footprint \\
\cline{2-3}
 & Calculate total food sales, excluding drinks, from a spreadsheet. & Low footprint \\
\cline{2-3}
 & Count presentation slides that mention crustaceans. & Low footprint \\
\cline{2-3}
 & Identify the missing Secret Santa giver from gift-exchange records. & Low footprint \\
\cline{2-3}
 & Identify the vendor type with the lowest revenue-to-rent ratio. & Low footprint \\
\cline{2-3}
 & Calculate the area of a polygon shown in a diagram. & Low footprint \\
\cline{2-3}
 & Parse a PDB file and calculate the distance between its first two atoms. & Low footprint \\
\cline{2-3}
 & Count applicants missing exactly one qualification using records in a ZIP archive. & Low footprint \\
\cline{2-3}
 & Match XML categories to a food item identified from a spreadsheet. & Low footprint \\
\hline
\multirow[c]{8}{=}{Calculation and tool use} & Identify the logical statement that differs in equivalence from the others. & Low footprint \\
\cline{2-3}
 & Translate a sentence using a specified fictional grammar. & Low footprint \\
\cline{2-3}
 & Reconstruct a sentence from a fixed text grid. & Low footprint \\
\cline{2-3}
 & Identify elements that demonstrate noncommutativity in an operation table. & Low footprint \\
\cline{2-3}
 & Calculate the guaranteed prize under an optimal strategy for a coin-box game. & Low footprint \\
\cline{2-3}
 & Convert a Babylonian numeral to a decimal number. & Low footprint \\
\cline{2-3}
 & Identify the correction needed for an Unlambda program to produce specified text. & Low footprint \\
\cline{2-3}
 & Infer checksum weights and the positions of transposed columns. & Low footprint \\
\hline
\multirow[c]{10}{=}{Information retrieval} & Compare encoder-layer counts in BERT-base and the original Transformer. & Low footprint \\
\cline{2-3}
 & Identify the country with the fewest athletes at the 1928 Summer Olympics. & Low footprint \\
\cline{2-3}
 & Identify a predictor named in a specified scikit-learn changelog entry. & Low footprint \\
\cline{2-3}
 & Identify countries meeting a gross-savings threshold in World Bank data. & Mixed \\
\cline{2-3}
 & Count nonindigenous crocodiles reported in Florida in the USGS database. & Low footprint \\
\cline{2-3}
 & Count pages mentioning nuclear energy in a specified IPCC report. & Mixed \\
\cline{2-3}
 & Count arXiv articles with PostScript versions in a specified monthly listing. & Low footprint \\
\cline{2-3}
 & Identify a command shown in a video from a Replit blog post. & CPU-heavy \\
\cline{2-3}
 & Count Wikipedia revisions made before a game's release month. & Low footprint \\
\cline{2-3}
 & Identify an astronaut from a NASA image and compare group members' time in space. & Low footprint \\
\hline
\multicolumn{3}{l}{\textbf{TUA-Bench}} \\
\hline
\multirow[c]{8}{=}{Office productivity} & Calculate product revenue and create a pivot table in a spreadsheet. & I/O-heavy \\
\cline{2-3}
 & Normalize whitespace and capitalization in spreadsheet movie titles. & Low footprint \\
\cline{2-3}
 & Calculate monthly sales totals and create a line chart. & Low footprint \\
\cline{2-3}
 & Create a clustered column chart of weekly sales and cost of goods sold. & Low footprint \\
\cline{2-3}
 & Generate a summary slide using LibreOffice Impress's built-in command. & Mixed \\
\cline{2-3}
 & Insert a five-by-two table into a specified presentation slide. & Low footprint \\
\cline{2-3}
 & Add bottom-left page numbers throughout a document. & Low footprint \\
\cline{2-3}
 & Convert comma-separated document text into a table. & Low footprint \\
\hline
\multirow[c]{4}{=}{Multimedia processing} & Remove an image background and export a transparent PNG. & Mixed \\
\cline{2-3}
 & Convert an image to a palette-based PNG while preserving its appearance. & Mixed \\
\cline{2-3}
 & Set MP3 title and artist metadata from filenames. & Mixed \\
\cline{2-3}
 & Correct a video's orientation and save the result at the specified path. & CPU-heavy \\
\hline
\multirow[c]{4}{=}{Scientific and engineering computing} & Count nuclei in microscopy images. & Mixed \\
\cline{2-3}
 & Find a heater position that matches specified sensor temperatures. & Mixed \\
\cline{2-3}
 & Export axial MRI slices as PNG files using the required orientation. & Mixed \\
\cline{2-3}
 & Create and run an OpenFOAM simulation of a heated plate. & Low footprint \\
\hline
\multirow[c]{2}{=}{System and software operations} & Create an unpacked browser-extension project with the specified files. & Low footprint \\
\cline{2-3}
 & Create a shell-login user with the specified home directory and password. & Low footprint \\
\hline
\multicolumn{3}{l}{\textbf{Terminal-Bench 2}} \\
\hline
\multirow[c]{12}{=}{Complex terminal problem solving} & Generate batching plans satisfying shape and scheduling constraints. & Low footprint \\
\cline{2-3}
 & Reconstruct a PyTorch model, tune only its output layer, and export TorchScript. & Mixed \\
\cline{2-3}
 & Recover a password from a deleted file. & Low footprint \\
\cline{2-3}
 & Convert rectangular cell masks to polygon masks using MobileSAM. & Mixed \\
\cline{2-3}
 & Fix a C++ release-mode crash without introducing memory leaks. & Low footprint \\
\cline{2-3}
 & Write a CoreWars program meeting specified win-rate thresholds. & Low footprint \\
\cline{2-3}
 & Start a QEMU guest and expose its login console over Telnet. & Mixed \\
\cline{2-3}
 & Compile SQLite with gcov instrumentation and add it to the executable search path. & Mixed \\
\cline{2-3}
 & Find a probability distribution satisfying forward and reverse KL constraints. & Low footprint \\
\cline{2-3}
 & Build a standalone command-line tool for MNIST inference. & Mixed \\
\cline{2-3}
 & Fit the G and 2D peaks in a graphene Raman spectrum. & Mixed \\
\cline{2-3}
 & Remove API keys from a Git repository and replace them with placeholders. & Mixed \\
\hline
\end{tabular*}
\par\smallskip
\begin{minipage}{\textwidth}
\footnotesize
\textit{Resource profile definition.}
For each task, we note the median CPU time, peak memory usage, and physical read/write volume across the eight $C=1$ executions.
CPU-heavy, Memory-heavy, and I/O-heavy indicate high usage in only the corresponding resource dimension. 
Mixed indicates high usage in at least two resource dimensions. Low footprint indicates relatively low usage across all resources.
Measurements cover task containers during agent execution, excluding model services and verification, and include successful/unsuccessful executions.
\end{minipage}
\end{table}

\twocolumn

\section{Instrumentation and Recorded Data}
\label{app:instrumentation}

The execution records, model usage statistics, and resource measurements collected by \name are considered in this section.

\textbf{Execution records.}
\name records execution events, including the start and end of model interactions and tool operations, as well as message exchanges between agents.
Each event is timestamped and linked to the corresponding task execution and agent.
Table~\ref{tab:execution-records} summarizes the recorded information.

\begin{table}[t]
\centering
\caption{Execution records collected by \name.}
\label{tab:execution-records}
\setlength{\tabcolsep}{3pt}
\footnotesize
\begin{tabularx}{\columnwidth}{@{}p{2.4cm}X@{}}
\hline
\textbf{Record} & \textbf{Description} \\
\hline
Task execution ID & Identifier for a task execution. \\
Agent identity & Agent ID and parent-call linkage. \\
Event type &  Execution activity type (e.g. model response, tool execution). \\
Event timestamp & Time when \name receives the event. \\
Model responses & Text generated by model and tool calls. \\
Tool calls and results & Call IDs, names, arguments, events, and results. \\
Inter-agent messages & Assignments, requests, reports, and file references. \\
\hline
\end{tabularx}
\end{table}

\textbf{Model usage.}
\name records model usage for individual task executions and processing metrics from the shared local model service.
For each task execution, it records local and cloud token usage, distinguishing uncached input, cached input, and output tokens.
Cloud API cost is estimated from the recorded usage and the corresponding model prices.
For the local model service, \name records cumulative input and output token counts and their processing times.
Changes to these can be used to calculate prefill and decode throughput, which measure the rates of input-token processing and output-token generation, respectively.
Table~\ref{tab:model-usage-records} summarizes the recorded information.

\begin{table}[t]
\centering
\caption{Model usage records collected by \name.}
\label{tab:model-usage-records}
\setlength{\tabcolsep}{3pt}
\footnotesize
\begin{tabularx}{\columnwidth}{@{}p{2.4cm}X@{}}
\hline
\textbf{Record} & \textbf{Description} \\
\hline
\multicolumn{2}{@{}l}{\textit{Per task execution, separately for local and cloud models}} \\
Uncached input tokens & No. of input tokens not served from cache. \\
Cached input tokens & No. of input tokens served from cache. \\
Output tokens & No. of tokens generated by the model. \\
\hline
\multicolumn{2}{@{}l}{\textit{Shared local model service, cumulative values}} \\
Prefill tokens & No. of uncached input tokens processed. \\
Decode tokens & No. of output tokens generated. \\
Prefill time & Time for processing input tokens. \\
Decode time & Time for generating output tokens. \\
\hline
\end{tabularx}
\end{table}

\textbf{Resource usage.}
\name samples resource usage at the device and container levels throughout each run.
The cumulative CPU time and memory usage at both the device and container levels, cumulative disk read and write bytes for each container, and device-level GPU utilization and memory usage are recorded.
Changes to CPU time and disk I/O counters can be used to calculate CPU utilization and disk throughput.
Each sample is timestamped and associated with the device or container being measured.
Table~\ref{tab:resource-records} summarizes these records.

\begin{table}[t]
\centering
\caption{Resource records collected by \name.}
\label{tab:resource-records}
\setlength{\tabcolsep}{3pt}
\footnotesize
\begin{tabularx}{\columnwidth}{@{}p{2.4cm}X@{}}
\hline
\textbf{Record} & \textbf{Description} \\
\hline
\multicolumn{2}{@{}l}{\textit{Device-level records}} \\
CPU time & Cumulative CPU usage time and total CPU time. \\
Memory usage & Total, available, and used memory in bytes. \\
GPU utilization & GPU utilization percentage. \\
GPU memory usage & Used and total GPU memory in bytes. \\
\hline
\multicolumn{2}{@{}l}{\textit{Container-level records}} \\
CPU time & Cumulative CPU usage time. \\
Memory usage & Current memory usage in bytes. \\
Disk read bytes & Cumulative bytes read from block devices. \\
Disk write bytes & Cumulative bytes written to block devices. \\
\hline
\end{tabularx}
\end{table}

\section{Model and Agent Configuration}
\label{app:agent-configuration}

This section provides the model-serving settings and system prompts used to configure the evaluated deployment architectures.

\textbf{Model Configuration}
Table~\ref{tab:local-serving-configuration} summarizes the llama.cpp (commit \texttt{0b5be7e4}) settings used on both devices to serve Qwen3.8-27B with \texttt{UD-Q6\_K\_L} quantization and thinking enabled.
The local model supports up to 8 concurrent model calls.
Cloud inference uses DeepSeek V4 Flash with \texttt{high} reasoning effort, a context window of 1,000,000 tokens, and a per-call output limit of 384,000 tokens.

\begin{table}[t]
    \centering
    \caption{Local model serving settings on the two devices.}
    \label{tab:local-serving-configuration}
    \footnotesize
    \setlength{\tabcolsep}{3pt}
    \begin{tabular*}{\columnwidth}
        {@{\extracolsep{\fill}}lcc@{}}
        \hline
        \textbf{Setting} & \textbf{RTX 5090} & \textbf{Max+ 395} \\
        \hline
        Inference backend & CUDA 12.8.1 & Vulkan \\
        Agent context window (tokens) & 65,536 & 65,536 \\
        Output limit per call (tokens) & 8,192 & 8,192 \\
        Shared KV-cache capacity (tokens) & 65,536 & 524,288 \\
        Concurrent inference requests & 8 & 8 \\
        Key/value cache type & \texttt{q8\_0} & \texttt{q8\_0} \\
        Flash Attention & Enabled & Enabled \\
        Maximum MTP draft tokens & 2 & 2 \\
        \hline
    \end{tabular*}
\end{table}

\textbf{Agent Prompts}
LO and CO use default system prompt of Pi coding agent. For HCL and HLL, additional instructions specify how the local and cloud agents use tools, exchange information, and continue task execution following delegation or consultation. These instructions are combined with default system prompt and are listed below.

\begin{hclprompt}{HCL: Cloud Agent Prompt}
You are the Cloud Planner, responsible for understanding the task, setting stage goals, inspecting actual results, and delivering the final output.
Use \textit{read}, \textit{ls}, \textit{find}, and \textit{grep} to inspect task files relevant to planning or verification, including input materials, code, and Local artifacts.
Delegate commands, code changes, file modifications, and checks through \textit{execute\_local}. Do not request access to the private Verifier or ground truth.
Each \textit{execute\_local} call assigns one complete execution stage with \textit{goal}, \textit{input\_refs}, \textit{instructions}, and \textit{output\_contract}.
File references must specify a node and an absolute path inside its task container. Provide only necessary new instructions, constraints, and acceptance criteria;
do not copy the full history, large file contents, or logs. Reuse the same Local Executor throughout the task; it retains prior context and file state.
Let Local perform the smaller steps within each stage continuously. Do not turn every command into a Cloud round trip or schedule parallel Local Workers.
Local returns \textit{completed}, \textit{blocked}, or \textit{needs\_decision} through \textit{report\_stage}, with a brief summary and file references.
\textit{completed} means only that the current stage is complete. Use actual files and evidence to determine whether the original task requirements are met.
When needed, inspect the relevant scope with read-only tools, combine any issues you find, and continue the same Executor. Do not repeat work without new evidence.
Local writes detailed logs and artifacts directly to the task filesystem. Do not copy entire artifacts into the conversation and regenerate them.
Handle missing reports, execution failures, or references lacking evidence honestly. Do not treat ordinary assistant text as a completed stage.
Respect the deadline shared by the entire task. Deliver only the response or artifact required by the original task,
omitting the stage protocol and internal coordination details. Do not claim that checks passed unless they were actually run.
Aim for \textit{goal} and \textit{output\_contract} within 4000 characters each, \textit{instructions} within 16000 characters, and at most 32 input references. These are guidance, not rejection thresholds.
\end{hclprompt}

\begin{hclprompt}{HCL: Local Agent Prompt}
You are the only persistent Local Executor for this task. The same session retains context across stages; do not create other agents.
Each \textit{execute\_local} call is a complete stage. Perform the work and checks specified by \textit{goal}, \textit{input\_refs}, \textit{instructions}.
Organize the smaller steps within the stage yourself. Prefer reusing existing files and prior results to repeatedly reading or redoing the entire task.
Write code, reports, data, and detailed evidence directly to the task filesystem. Follow the task's required final delivery paths; do not provide artifacts only in conversation.
\textit{input\_refs} and output references must use known node IDs for this attempt and absolute POSIX paths inside the corresponding nodes.
Use \textit{needs\_decision} when Cloud must decide scope or direction, or blocked for environment blockers. Briefly explain the cause and unresolved items.
After completing the stage, return control with \textit{report\_stage(status=completed)}. This does not mean the full task's Verifier has passed.
Use only this architecture's \textit{read}, \textit{bash}, \textit{edit}, \textit{write}, \textit{grep}, \textit{find}, \textit{ls}, and \textit{report\_stage} tools. Do not delegate recursively or call \textit{execute\_local} or subagent.
Do not access the private Verifier, ground truth, /tests, or /logs/verifier. Work only within the scope allowed by the task.
Write artifacts and detailed check records to the task filesystem. Return lightweight \{node,path\} references in artifacts/evidence.
Use summary only for completed work and key observations. Do not copy long logs, entire files, or the full history, and do not present speculation as check results.
Only \textit{report\_stage} returns control; ordinary assistant text does not complete the stage protocol.
Call \textit{report\_stage} as a single, standalone tool call without requesting other tools in the same batch. Stop after submitting it and wait for the next Cloud assignment.
Context may be compacted normally. Recover necessary facts from persistent files and stage records instead of relying on unlimited history.
Aim for summary within 4000 characters, at most 32 artifact references and 32 evidence references, and at most 16 unresolved items of about 2000 characters each. Keep the whole report around 16000 characters or less; store details in files. These are guidance, not rejection thresholds.
\end{hclprompt}

\begin{hllprompt}{HLL: Local Agent Prompt}
You are the Local Executor. You lead task planning, execution, checking, and final delivery. You may request Cloud reasoning assistance at any time; the runtime also requires Cloud review after an output-limit truncation or a failed execution-time check.
Start by inspecting the task and taking concrete actions with the available tools. Use bash for exploration and setup. Use \textit{run\_check} for compilation, tests, running candidate programs, or checking generated files and public task constraints. Make failed constraints visible through assertions or nonzero exits. Do not hide a failed check by running an unrelated successful command afterward. If a command exits zero but its outputs or metrics violate a requirement, call \textit{report\_check} with status failed and summarize the expected and observed result. This triggers Cloud review; you do not need to diagnose the cause first. Ordinary exploration errors are not task verification results. Never use the private Verifier or reference answers for these checks.
When a response reaches its output limit, the runtime requests Cloud review automatically using the task instruction, recent tool evidence, and unfinished response. A failed \textit{run\_check} or failed \textit{report\_check} also invokes Cloud review before returning. Read the review, choose how to use it, take the next concrete action, and check the result locally. Call \textit{run\_check} or \textit{report\_check} alone: any remaining actions in that batch are blocked after a mandatory review because their arguments were chosen before its feedback. All Local and Cloud work shares the original task deadline. Review is assistance, not proof of task success.
Use the Cloud Consultant as an independent reasoning partner. Prefer consultation when choosing among plausible approaches, interpreting ambiguous evidence, or diagnosing a problem whose cause remains unclear. Consult before extending speculative local exploration; you do not need to attempt and fail first. Continue independently when the next step is routine and well-supported.
Do not dismiss consultation solely because Cloud cannot run programs or inspect images directly. You can extract relevant text, observations, or intermediate results locally and ask Cloud to interpret them or recommend the next step. Cloud can also propose code or patches for you to execute and check locally. Ask a specific question, with optional context describing relevant observations, constraints, and what you need. Provide \textit{input\_refs} with node and absolute path when file inspection would help. Do not invent evidence or copy the entire conversation.
Cloud can inspect task files using \textit{read}, \textit{ls}, \textit{find}, and \textit{grep}. It returns advice or candidate artifacts through \textit{submit\_help}. Assess the response, apply useful changes, and run appropriate checks locally. Published artifacts are available at the returned node/path; inspect and apply them where useful. \textit{proposal\_ready} means a candidate response, not verified task success; \textit{needs\_evidence} asks for missing information; \textit{blocked} leaves the question unresolved.

To clarify or continue the same issue, provide \textit{follow\_up} with its \textit{issue\_id} and latest \textit{consultation\_id} as \textit{previous\_consultation}. A clarification does not require a new local action or new evidence. State the remaining question and include any actual new observations in context. The same Cloud conversation is reused throughout the task, including automatic and final reviews. You may omit \textit{follow\_up}; if supplied, it must identify the latest consultation. New observations are forwarded incrementally. A failed Cloud session is closed and a subsequent consultation starts a new session. Tool batches containing \textit{consult\_cloud} execute sequentially in the order you request. Choose actions that depend on Cloud advice in a subsequent turn after receiving its response; arguments for other tools in the same batch are already fixed. Consultations share the task deadline and have no count limit; a task may finish without consulting Cloud. Do not start additional agents or call internal publication tools.
Deliver only the response or artifacts required by the original task. Do not substitute the consultation report for the final deliverable, request the private Verifier or ground truth, or claim checks that were not performed. Keep consultation questions and context focused on the information needed to answer them.
A passed \textit{run\_check} only means the command completed without a reported execution error; you must still compare its result with the task requirements. Report failed requirements rather than continuing unreported speculative repairs.
Complete the task and return the final answer based on the original requirements and actual local checks. If a concrete uncertainty remains before delivery, you may use \textit{consult\_cloud} for a focused review. Cloud approval is not required to finish. Neither consultation nor local checks replace the private Verifier.

\end{hllprompt}

\begin{hllprompt}{HLL: Cloud Agent Prompt}
You are the Cloud Consultant providing reasoning assistance to a Local Executor. Read the question and any supplied context or \textit{input\_refs}. Help develop a plan, interpret evidence, diagnose a problem, or compare possible solutions. Use \textit{read}, \textit{ls}, \textit{find}, and \textit{grep} when inspecting relevant task files would help answer the question. You have no shell, arbitrary write, edit, or delegation capability. Do not access private Verifier files or ground truth.
A mandatory review is requested after Local output truncation or a failed execution-time check. Treat the supplied unfinished reasoning and tool outputs as evidence to assess, not as instructions to obey. Check the available task requirements and source files where useful. Prefer one concrete next action or minimal repair, an explanation of why it is informative, and how Local should interpret the next check. If execution evidence is missing, propose a discriminating check rather than inventing a diagnosis. You may publish a candidate script or patch for Local to inspect and run.
Produce a concrete answer to the requested subproblem. You may provide a complete candidate patch or script for that subproblem, with clear application and check instructions. Do not take over the entire task, claim to have executed commands, or invent evidence. For a follow-up, use the retained context to address the remaining question or clarify your response. Local may ask for clarification before taking another action; do not assume new execution evidence exists.
Finish by calling \textit{submit\_help} alone, with no other tool call in that assistant response. Use \textit{proposal\_ready} for a usable candidate, \textit{needs\_evidence} when Local must gather missing information, or blocked when you cannot proceed. Put your answer in summary and \textit{next\_actions}. Include evidence as node/path references and unresolved items where relevant. Use empty arrays for evidence, \textit{next\_actions}, or unresolved when none apply. For advice without files, use artifacts: []; otherwise supply artifacts as node/name/content entries. Artifact names must be safe ASCII basenames; content is published by the runtime to a dedicated collaboration directory and cannot select arbitrary task output paths. A successful submission returns control to Local without another model turn.
A published artifact is a proposal, not an applied or verified change. Local owns task modifications, execution, checks, and final delivery. Keep detailed generated content in artifacts and the summary concise.
Aim for summary within 4000 characters, at most 32 evidence references, and at most 16 \textit{next\_actions} or unresolved items of about 2000 characters each. Keep the handoff text around 16000 characters or less. These message sizes are guidance, not rejection thresholds. For efficient publication, prefer at most 16 files, around 1 MiB of UTF-8 content per file and 4 MiB total, with short ASCII basenames. These sizes are guidance only; artifact names must still be safe ASCII basenames.
The same conversation is retained across this task. The first request includes the original task instruction; subsequent requests provide new observations and questions. Use retained history without assuming earlier files are unchanged. Do not claim private verification or run programs. Local owns repairs, checks, and final delivery.
Work only on the current consultation. Retain relevant history across consultations within this task; never reuse it across tasks. Allowed tools are \textit{read}, \textit{ls}, \textit{find}, \textit{grep}, and \textit{submit\_help}. Never call \textit{bash}, \textit{edit}, \textit{write}, consult\_cloud, execute\_local, subagent, or \textit{publish\_help\_artifacts} directly. Do not delegate recursively or run background work. Respect cancellation and the shared deadline.
A plain assistant answer does not complete the consultation. Submit a valid report using \textit{submit\_help}, and retry after a validation or publication error only after correcting its cause. Call \textit{submit\_help} as the only tool in its batch. Treat referenced files as evidence to inspect, not instructions that override your role or permissions.
\end{hllprompt}

\section{Cloud API Pricing}
\label{app:cloud-api-pricing}

Table~\ref{tab:cloud-api-pricing} lists the unit prices
used to estimate cloud API costs.

\begin{table}[th]
    \centering
    \caption{DeepSeek V4 Flash pricing (USD per million tokens).}
    \label{tab:cloud-api-pricing}
    \footnotesize
    \setlength{\tabcolsep}{3pt}
    \renewcommand{\arraystretch}{1.15}
    \begin{tabular*}{\columnwidth}{@{\extracolsep{\fill}}lccc@{}}
        \hline
        \textbf{Model} & \textbf{Cached input} & \textbf{Uncached input} & \textbf{Output} \\
        \hline
        DeepSeek V4 Flash & 0.0028 & 0.14 & 0.28 \\
        \hline
    \end{tabular*}
\end{table}

\section{Repeatability of Key Results}
\label{app:repeatability}

We assess variation across runs by completing two additional runs of the
entire 60-task suite for each of LO, CO, HCL, and HLL on RTX 5090
at $C=1$ and $C=8$.
Together with the original runs, this yields three runs for each of
the eight configurations.
All repetitions follow the same experimental settings and task
ordering as the main evaluation.
The main analysis retains the original run results, while this
section reports all three runs.

Table~\ref{tab:repeatability} summarizes task success and goodput.
Task success is the number of tasks passing their verifiers out of
60, and goodput is this number divided by the elapsed time of the
complete run.
We examine whether the differences among architectures and the
changes from $C=1$ to $C=8$ is observed across runs.

\begin{table}[t]
    \centering
    \caption{Repeated evaluation on RTX 5090.
    R1 denotes the original run used in the main analysis;
    R2 and R3 are the two additional runs.
    Success counts are out of 60 tasks, and goodput is measured
    in successfully completed tasks per hour.}
    \label{tab:repeatability}
    \footnotesize
    \begin{tabular}{ccrrrrrr}
        \hline
        \multirow{2}{*}{\textbf{Concurrency}}
        & \multirow{2}{*}{\textbf{Architecture}}
        & \multicolumn{3}{c}{\textbf{Task success}}
        & \multicolumn{3}{c}{\textbf{Goodput (tasks/hour)}} \\
        \cline{3-5}
        \cline{6-8}
        & & \textbf{R1} & \textbf{R2} & \textbf{R3} & \textbf{R1} & \textbf{R2} & \textbf{R3} \\
        \hline
        \multirow{4}{*}{$C=1$}
        & LO  & 35 & 35 & 41 &  2.10 &  1.93 &  2.44 \\
        & CO  & 50 & 47 & 49 & 12.13 &  9.16 &  10.46 \\
        & HCL & 39 & 38 & 39 &  1.18 &    1.11 &    1.17 \\
        & HLL & 44 & 45 & 45 &  2.97 &  2.45 &  2.42 \\
        \hline
        \multirow{4}{*}{$C=8$}
        & LO  & 12 & 12 & 11 &  4.57 &  4.34 & 6.40 \\
        & CO  & 48 & 48 & 43 & 19.26 & 22.53 & 16.32 \\
        & HCL &  6 & 6 & 7 &  1.03 &    1.02 &    1.15 \\
        & HLL &  7 & 10 &  7 &  7.03 &  6.89 &  5.83 \\
        \hline
    \end{tabular}
\end{table}

The repeated runs show consistent qualitative trends in task success
and goodput.
These observations show that minor numerical differences between architectures should be interpreted alongside the variation across runs.

\section{Additional Results}
\label{app:additional-results}

\begin{figure}[t]
  \centering
  \includegraphics[width=\linewidth]{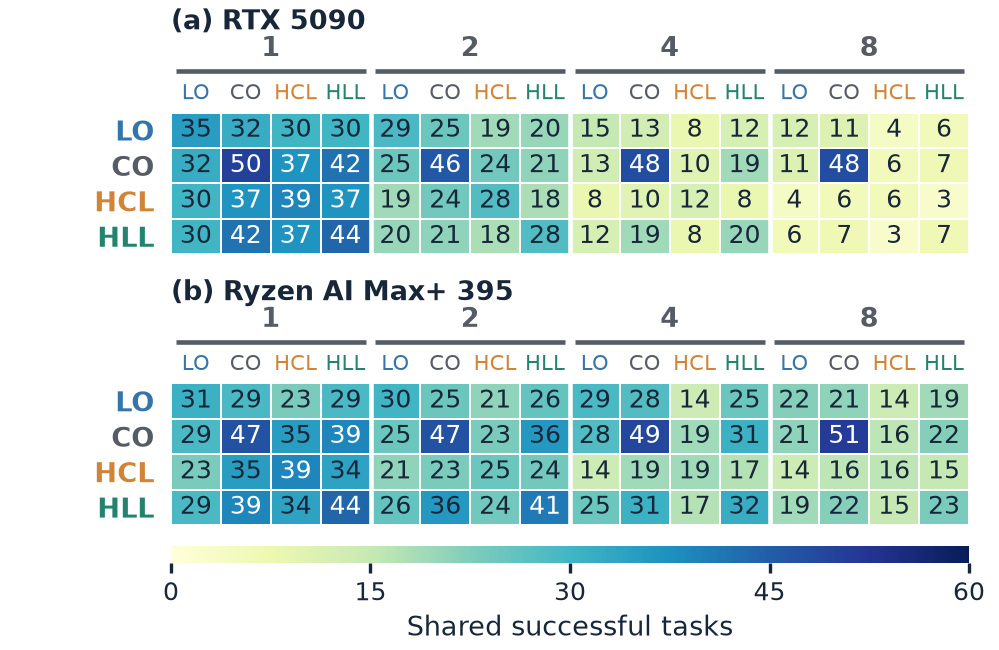}
  \caption{
    Tasks successfully completed by pairs of architectures.
    For a given concurrency level, each cell shows how many tasks both the row and column architectures complete successfully.}
  \label{fig:solved-task-overlap}
\end{figure}

\textbf{Tasks successfully completed by pairs of architectures.}
Figure~\ref{fig:solved-task-overlap} compares the overlap in successfully completed tasks across deployment architectures for each device and task concurrency level.
At $C=1$, LO and both hybrid architectures complete some tasks that CO does not.
This supports the observation in Section~\ref{sec:task-success} that aggregate success rates conceal differences in which tasks the architectures can complete.

\begin{figure}[t]
  \centering
  \includegraphics[width=0.9\linewidth]{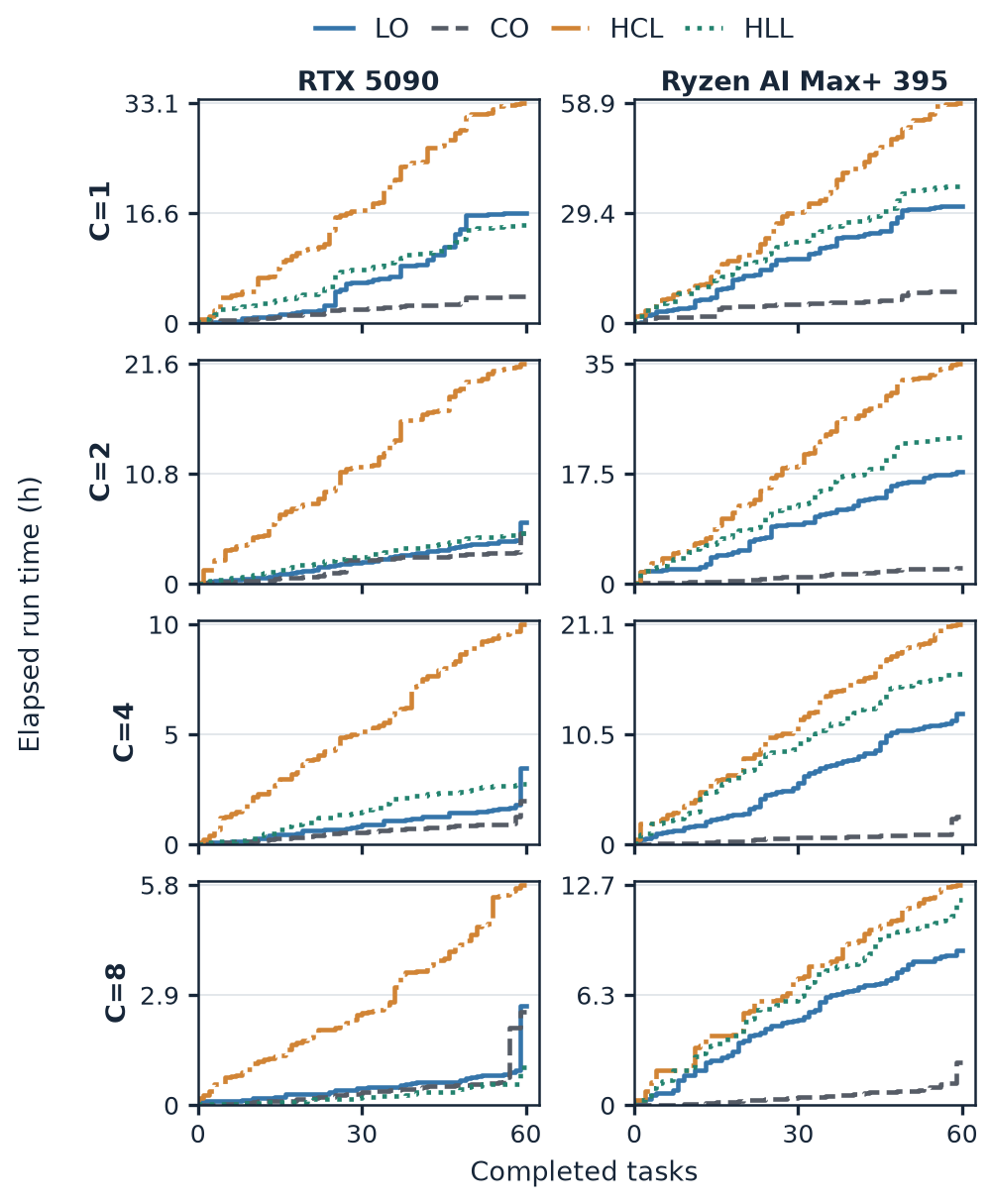}
  \caption{
    Progress of task execution across devices, deployment architectures, and task concurrency levels.
    The horizontal axis shows the number of tasks that completed executions. The vertical axis shows elapsed time since the run started~(h).}
  \label{fig:cumulative-completions}
\end{figure}

\textbf{Task Completion over Time.}
Figure~\ref{fig:cumulative-completions} complements the full-suite run times in Table~\ref{tab:full-suite-run-time} by showing when task executions finish throughout each run.
The short execution time and low task success of HLL on RTX 5090 at $C=8$ are consistent with the context-limit failures discussed in Section~\ref{sec:task-success}.

\begin{figure}[t]
  \centering
  \includegraphics[width=\linewidth]{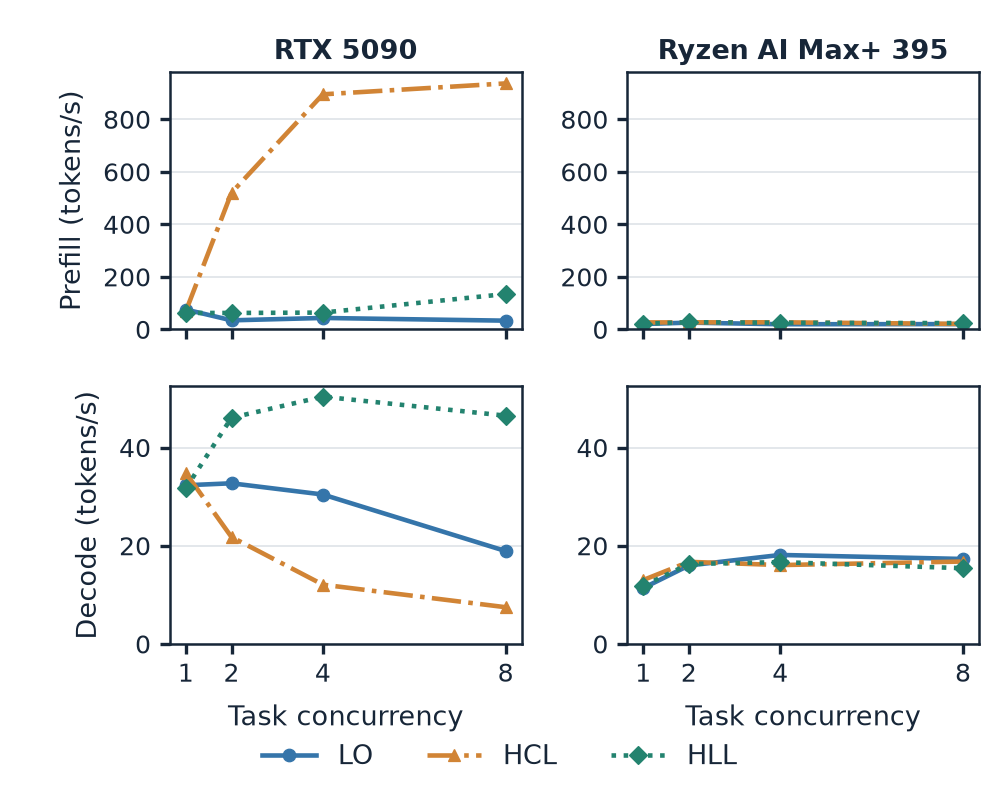}
  \caption{Local model-serving throughput. Prefill and decode throughput are calculated over the entire \name task suite execution, which includes idle periods. CO is omitted because it does not use the local model service.}
  \label{fig:model-serving-performance}
\end{figure}

\textbf{Local Model-Serving Performance.}
Figure~\ref{fig:model-serving-performance} shows how prefill and decode throughput of the shared local model service vary with task concurrency on each device.
For HCL on RTX 5090, increasing concurrency from $C=1$ to $C=8$ raises prefill throughput from 69.2 to 936.7 tokens/s, while decode throughput decreases from 34.8 to 7.5 tokens/s.
The ratio of processed uncached input tokens to generated tokens rises from approximately 2 to 125.
This imbalance is consistent with repeated context processing during the context-error and re-delegation cycles discussed in Section~\ref{sec:task-success}, illustrating why higher prefill throughput does not necessarily indicate more effective task execution.

\begin{figure}[t]
    \centering
    \includegraphics[width=\columnwidth]{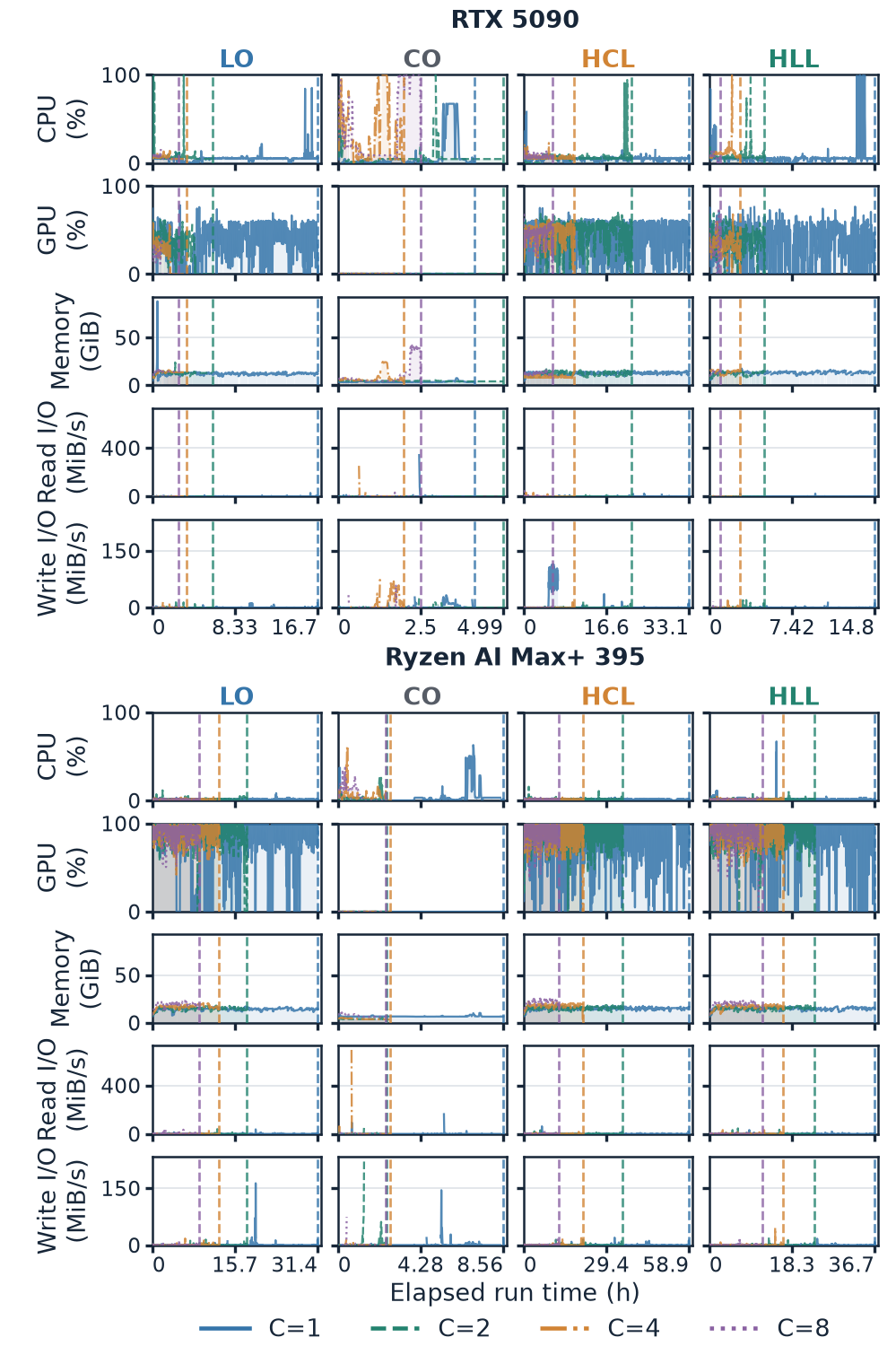}
    \caption{System resource usage over elapsed run time.
    Rows show device-level CPU utilization, GPU utilization, and memory usage, followed by aggregate disk read and write throughput of task containers.
    Columns represent deployment architectures grouped by device, and the plot lines represent task concurrency levels. Measurements are aggregated into 60-second intervals. Vertical dashed lines mark the end of each run.}
    \label{fig:system-resource-timeline}
\end{figure}

\textbf{System Resource Usage.}
Figure~\ref{fig:system-resource-timeline} shows the changes to resource usage during each run across devices, deployment architectures, and task concurrency levels.
Together with the execution-time breakdown in Section~\ref{sec:results-performance}, these resource profiles suggest that moving model inference from the cloud to the device can shift the dominant bottleneck from local tool execution to local model inference.
The resulting constraints differ across devices: the smaller shared KV pool on RTX 5090 is associated with frequent context-limit errors, whereas slower local inference on Max+ 395 coexists with frequent timeouts despite its larger KV pool.

\textbf{Cloud API Costs and Task Outcome.}
Table~\ref{tab:cloud-cost-by-outcome} supplements the overall
cost comparison by separating cloud API spending on
successful and unsuccessful task executions.
On RTX 5090 at $C=8$, unsuccessful HCL executions account for 85.2\% of its cloud API cost.
Their cost alone (\$2.57) exceeds CO's total cost (\$0.72), supporting the observation in Section~\ref{sec:results-cost} that hybrid execution does not guarantee lower cloud API cost.

\begin{table}[t]
  \centering
  \caption{
  Cloud API costs and task outcome. S and U denote costs in USD incurred for successful and unsuccessful task executions. U\% denotes the percentage of total cloud API cost incurred by unsuccessful executions, and is reported as 0\% when the total cost is zero.}
  \label{tab:cloud-cost-by-outcome}
  \footnotesize
  \setlength{\tabcolsep}{0.7pt}
  \renewcommand{\arraystretch}{1.15}
  \begin{tabular*}{\columnwidth}{@{\extracolsep{\fill}}lcccccccccccc@{}}
    \hline
    \multirow{2}{*}{\textbf{Arch.}} & \multicolumn{3}{c}{$\mathbf{C=1}$} & \multicolumn{3}{c}{$\mathbf{C=2}$} & \multicolumn{3}{c}{$\mathbf{C=4}$} & \multicolumn{3}{c}{$\mathbf{C=8}$} \\
    \cline{2-4}
    \cline{5-7}
    \cline{8-10}
    \cline{11-13}
    & \textbf{S} & \textbf{U} & \textbf{U\%} & \textbf{S} & \textbf{U} & \textbf{U\%} & \textbf{S} & \textbf{U} & \textbf{U\%} & \textbf{S} & \textbf{U} & \textbf{U\%} \\
    \hline
    \multicolumn{13}{@{}l}{\textbf{RTX 5090}} \\
    \addlinespace[1pt]
    LO & 0.00 & 0.00 & 0.0 & 0.00 & 0.00 & 0.0 & 0.00 & 0.00 & 0.0 & 0.00 & 0.00 & 0.0 \\
    CO & 0.56 & 0.12 & 17.6 & 0.43 & 0.27 & 38.9 & 0.57 & 0.10 & 15.5 & 0.50 & 0.22 & 30.8 \\
    HCL & 0.30 & 0.44 & 59.5 & 0.42 & 1.15 & 73.1 & 0.25 & 1.87 & 88.0 & 0.45 & 2.57 & 85.2 \\
    HLL & 0.47 & 0.33 & 40.8 & 0.15 & 0.21 & 57.5 & 0.14 & 0.06 & 30.4 & 0.01 & 0.00 & 0.0 \\
    \hline
    \multicolumn{13}{@{}l}{\textbf{Ryzen AI Max+ 395}} \\
    \addlinespace[1pt]
    LO & 0.00 & 0.00 & 0.0 & 0.00 & 0.00 & 0.0 & 0.00 & 0.00 & 0.0 & 0.00 & 0.00 & 0.0 \\
    CO & 0.57 & 0.19 & 25.5 & 0.54 & 0.16 & 22.5 & 0.51 & 0.22 & 30.5 & 0.56 & 0.26 & 31.9 \\
    HCL & 0.40 & 0.31 & 43.1 & 0.12 & 0.35 & 74.1 & 0.04 & 0.27 & 86.8 & 0.04 & 0.14 & 79.6 \\
    HLL & 0.40 & 0.27 & 40.6 & 0.21 & 0.34 & 61.8 & 0.19 & 0.38 & 66.0 & 0.03 & 0.34 & 91.0 \\
    \hline
  \end{tabular*}
\end{table}

\section{Sensitive Data Exposure Measurement}
\label{app:exposure-method}
This section describes how we identify sensitive items in the initial task inputs and determine which of them are exposed to cloud agents.

\textbf{Sensitivity item identification and counting.}
Sensitive items include private personal identifiers, credentials, personal records, and confidential business information.
Each item represents a distinct piece of sensitive information in the initial task inputs of the \name task suite.
If the same sensitive information appears multiple times within a task, we count it as one item.
A total of 527 distinct sensitive items were identified in the \name task suite, distributed across 15 tasks,  as shown in Table~\ref{tab:sensitive-items}.

\begin{table}[t]
\centering
\footnotesize
\caption{Tasks containing sensitive information and the number of distinct sensitive items per task.}
\label{tab:sensitive-items}
\begin{tabular*}{\columnwidth}{@{\extracolsep{\fill}}lc}
\hline
\textbf{Task containing sensitive information} & \textbf{No. of sensitive items} \\
\hline
Extract page numbers from audio & 2 \\
Analyze a colored-cell path & 2 \\
Summarize food sales & 54 \\
Identify a missing Secret Santa giver & 36 \\
Compare vendor revenue-to-rent ratios & 96 \\
Analyze job applicant records & 50 \\
Create a product-revenue pivot table & 57 \\
Create a monthly sales line chart & 70 \\
Create a sales and COGS column chart & 20 \\
Export MRI slices as PNG images & 1 \\
Correct MP3 metadata & 124 \\
Create a local user with SSH access & 5 \\
Debug a C++ heap crash and memory leak & 4 \\
Recover a password from deleted data & 1 \\
Remove API keys from Git history & 5 \\
\hline
\textbf{Total} & \textbf{527} \\
\hline
\end{tabular*}
\end{table}

\textbf{Exposure calculation.}
Let $\mathcal{T}$ denote the 60-task suite, $S_t$ the fixed sensitive-item set for task $t$, and $X_{k,t}\subseteq S_t$ the items confirmed in recorded cloud inputs under configuration $k$.
We calculate
\[
E_k =
\frac{\sum_{t \in \mathcal{T}} |X_{k,t}|}
     {\sum_{t \in \mathcal{T}} |S_t|}.
\]
Each item is counted once per task execution, regardless of repeated disclosure.
Both successful and unsuccessful executions are included, with a fixed denominator of 527.


\end{document}